\documentclass[sigconf]{acmart}

\AtBeginDocument{%
  \providecommand\BibTeX{{%
    \normalfont B\kern-0.5em{\scshape i\kern-0.25em b}\kern-0.8em\TeX}}}

\copyrightyear{2021}
\acmYear{2021}
\setcopyright{acmcopyright}\acmConference[KDD '21]{Proceedings of the 27th ACM SIGKDD Conference on Knowledge Discovery and Data Mining}{August 14--18, 2021}{Virtual Event, Singapore}
\acmBooktitle{Proceedings of the 27th ACM SIGKDD Conference on Knowledge Discovery and Data Mining (KDD '21), August 14--18, 2021, Virtual Event, Singapore} \acmPrice{15.00}
\acmDOI{10.1145/3447548.3467097}
\acmISBN{978-1-4503-8332-5/21/08}

\usepackage{hyperref}
\usepackage[ruled,vlined]{algorithm2e}
\usepackage{xcolor}
\usepackage{dsfont}
\usepackage[normalem]{ulem}
\usepackage{multirow}

\begin{document}
\fancyhead{}

\title{Device-Cloud Collaborative Learning for Recommendation}

\author{Jiangchao Yao*$^\dagger$, Feng Wang*$^\dagger$, Kunyang Jia*$^\dagger$, Bo Han$^{\ddagger}$, Jingren Zhou$^\dagger$, Hongxia Yang$^\dagger$}
\thanks{*These authors contributed equally to this research.}
\affiliation{
\institution{\textsuperscript{$\dagger$}DAMO Academy, Alibaba Group \city{Hang Zhou} \country{China}; 
\textsuperscript{$\ddagger$}Hong Kong Baptist University
  \city{Hong Kong}
  \country{China};} 
\textsuperscript{$\dagger$}\{jiangchao.yjc, wf135777, kunyang.jky, jingren.zhou, yang.yhx\}@alibaba-inc.com, \textsuperscript{$\ddagger$}bhanml@comp.hkbu.edu.hk
}


\begin{abstract}
  With the rapid development of storage and computing power on mobile devices, it becomes critical and popular to deploy models on devices to save onerous communication latencies and to capture real-time features. While quite a lot of works have explored to facilitate on-device learning and inference, most of them focus on dealing with response delay or privacy protection. Little has been done to model the collaboration between the device  and the cloud modeling and benefit both sides jointly. To bridge this gap, we are among the first attempts to study the Device-Cloud Collaborative Learning (DCCL) framework. Specifically, we propose a novel \emph{MetaPatch} learning approach on the device side to efficiently achieve ``thousands of people with thousands of models'' given a centralized cloud model. Then, with billions of updated personalized device models, we propose a ``model-over-models'' distillation algorithm, namely \emph{MoMoDistill}, to update the centralized cloud model. Our extensive experiments over a range of datasets with different settings demonstrate the effectiveness of such collaboration on both cloud and devices, especially its superiority to model long-tailed users.
\end{abstract}

\begin{CCSXML}
<ccs2012>
    <concept>
    <concept_id>10002951.10003317.10003347.10003350</concept_id>
    <concept_desc>Information systems~Recommender systems</concept_desc>
    <concept_significance>500</concept_significance>
    </concept>
</ccs2012>
\end{CCSXML}

\ccsdesc[500]{Information systems~Recommender systems}

\keywords{On-device Intelligence, Cloud Computing, Recommender Systems}

\maketitle

\section{Introduction}
Nascent applications for mobile computing and the Internet of Things (IoTs) are driving computing toward dispersion~\cite{satyanarayanan2017emergence}. Specially, the evolving capacity of mobile devices makes it possible to consider the intelligence services, e.g., online recommendation, from cloud to device modeling. Recently, several works in different perspectives like privacy~\cite{45648federated,karimireddy2019scaffold,bistritz2020distributed}, efficiency~\cite{han2016deep,cai2020tinytl}, applications~\cite{sundaramoorthy2018harnet,dai2019machine,gong2020edgerec} have explored this pervasive computing advantages. Mobile recommender systems with the on-device engines \textit{e.g.,} TFLite\footnote{\url{https://www.tensorflow.org/lite}} and CoreML\footnote{\url{https://developer.apple.com/documentation/coreml}} hence
attract more and more attention. 

Previous research in this area of recommender systems can be summarized into two lines, \emph{on-device inference} and \emph{on-device learning} given the centralized model. The former deploys the pretrained model to the devices, and executes the inference to save the onerous communication latencies and capture real-time features. This mainly solves the computational efficiency problem, yielding the exploration of the device-friendly model inference. For example, Sun \textit{et al.}~\cite{sun2020generic} proposed CpRec to shrink the sequential recommender system, which compress the size of input and output matrices as well as the parameters of middle layers via adaptive decomposition and parameter sharing. In a broader area, Cai \textit{et al.}~\cite{cai2020once} introduced a generalized network pruning method, progressive shrinking, to reduce the model size across dimensions for efficient deployment. Gong \textit{et al.}~\cite{gong2020edgerec} explored a split deployment across cloud and device to reduce the inference cost of on-device components.

The latter line of works aggregate the temporal on-device training pieces into a centralized model to overcome the privacy constraint, namely Federated recommender systems~\cite{yang2019federated}. The gradients of the local copy from the centralized deep model is first executed on plenty of devices, and then are collected to the server to update the model parameters by federated averaging (FedAvg) or its variants~\cite{45648federated,karimireddy2019scaffold,muhammad2020fedfast}. For example, Qi \textit{et al.}~\cite{qi2020privacy} applied Federated Learning and differential privacy for 
news recommendation and achieved a balance between recommendation performance and privacy protection. Lin \textit{et al.}~\cite{lin2020meta} introduced a MetaMF to account for the storage, energy and communication bandwidths in mobile environments, which distributes the gradient computation across cloud and devices. Niu~\textit{et al.}~\cite{niu2020billion} followed a similar split in gradient computation but designed a tunable privacy to the local submodel. 

\begin{figure*}[!ht]
    \centering
    \includegraphics[width=0.9\textwidth]{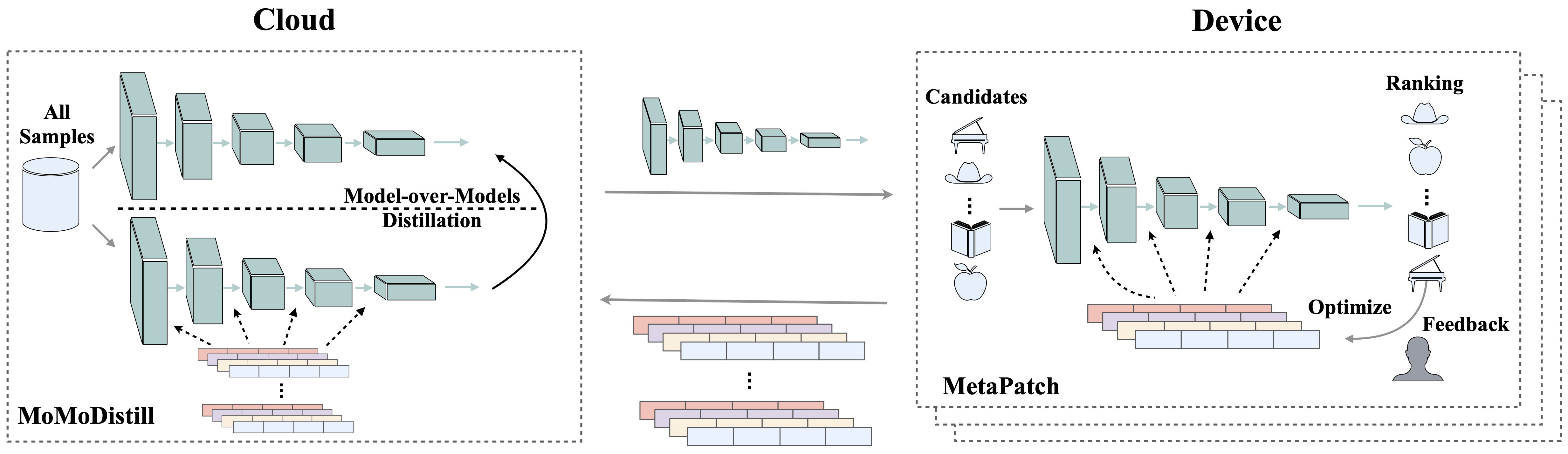}
    \caption{The general DCCL framework for recommendation. The cloud side is responsible to learn the centralized cloud model via the model-over-models distillation from the personalized on-device models. The device receives the centralized cloud model to conduct the on-device personalization. We propose \emph{MoMoDistill} and \emph{MetaPatch} to instantiate each side respectively.}
    \label{fig:model}
\end{figure*}
Different from the above two lines of works, we focus on how to leverage the advantages of the device modeling and the cloud modeling jointly to benefit both sides. The intuition behind this direction is two-fold: on the one hand, the centralized cloud model usually ignores or even sacrifices the experience of long-tailed samples, users or items, to maximize the global revenue. And long-tailed and non-i.i.d (independent and identically distributed) characteristics in recommendation samples inevitably induce the model bias to the minorities~\cite{park2008long}. One possible solution is to personalize the cloud model with local data on each device. On the other hand, training samples on each device alone may be quite limited suffering from local optimization~\cite{finn2017model}. In contrast, the centralized cloud models are updated with data from all devices and able to avoid this problem. This motivates us to propose a comprehensive Device-Cloud Collaborative Learning (DCCL) framework.

We summarize our contributions as follows:
\begin{itemize}
    \item Compared to existing works that either only consider the cloud modeling, or on-device inference, or the aggregation of the temporal on-device training pieces to handle the privacy constraint, we formally propose a Device-Cloud Collaborative Learning framework (DCCL) to benefit both sides.
    
    \item We propose two novel methods \emph{MetaPatch} and \emph{MoMoDistill} to instantiate each side in DCCL, which consider the sparsity challenge for on-device personalization, and enhance the centralized cloud model via the ``model-over-models" \emph{MoMoDistill}  distillation instead of the conventional ``model-over-data" paradigm.
    
    \item Extensive experiments on a range of datasets demonstrate that DCCL outperforms the state-of-the-art methods, and especially, we provide a comprehensive analysis about its advantage to long-tailed users and the inter-loops between computing interactions of cloud and device.
\end{itemize}
\section{Related Works}
\subsection{Recommender System}
Recommender systems~\cite{resnick1997recommender} have been widely studied in the last decade and become an indispensable infrastructure of web services in the era of cloud computing and big data. The related recommendation methods are gradually improved with the development of collaborative filtering, deep learning and sequential modeling. The early stage mainly focused on the user-based collaborative filtering~\cite{zhao2010user}, the item-based collaborative filtering~\cite{sarwar2001item} and matrix factorization~\cite{koren2009matrix,rao2016preference}. As deep learning achieved a great success in computer vision, several variants of collaborative 
filtering combined with deep neural networks were proposed~\cite{cheng2016wide,he2017neural,guo2017deepfm,cui2018variational,yao2017discovering,chen2020towards}. They leveraged the non-linear transformation of deep neural networks to activate high-level semantics for more accurate recommendation. Sequential modeling as another perspective to model the user interests has been successfully applied to recommender systems~\cite{shani2005mdp}. With the optimization in architectures, several methods based on GRU~\cite{jannach2017recurrent}, Attention~\cite{kang2018self,zhou2018deep,tan2021sparse,zhang2021cause} have achieved the remarkable performance in recommender systems.

\subsection{On-device Inference}
On-device inference as an important part of Edge AI~\cite{stoica2017berkeley} greatly reduces the onerous latency and incorporates rich features in recommendation. This line of works critically depend on the device capacity~\cite{bedi2018review}, efficient neural network architectures~\cite{han2015learning}, the model compression technique~\cite{han2016deep} and some split deployment strategies. Recently, several hardware efficient architectures like MobileNets~\cite{howard2017mobilenets} were proposed, which made the model size and the computational budget for on-device inference lightweight. Another effective perspective is the model compression based on the network pruning or quantization~\cite{han2016deep,cai2020once}, which reduces the units, the channels or the value accuracy of the parameters. In recommender systems, 
CpRec~\cite{sun2020generic} considered to construct a compressed counterpart of the sequential models by introducing a block-wise adaptive decomposition and parameter sharing scheme. Lee \textit{et al.}~\cite{lee2019device} described how to leverage the mobile GPU to run deep neural network tasks, which considers the special limited computing power, thermal constraints, and energy consumption. Dai \textit{et al.}~\cite{dai2019machine} presented an app to demonstrate the potential usage of on-device inference for mobile health applications. Some other works~\cite{gong2020edgerec} also explored the divide-and-conquer deployment to save the running time on devices, in which the computational prohibitive modules can be moved to the cloud. Nevertheless, all these works still share a centralized model.

\subsection{On-device Learning}
On-device learning is more time-consuming than on-device inference as it requires the extra computation regarding gradients~\cite{eshratifar2019jointdnn}. Current methods are mainly for Federated Learning~\cite{45648federated}, a distributed learning framework to keep the data in the local and only communicate gradients or parameters. It is important as more and more users consider the privacy protection on the Internet as General Data Protection Regulation (GDPR)\footnote{\url{https://gdpr-info.eu/}} claims in Europe. In Federated Learning, on-device learning is used to compute the temporal training pieces and send to the cloud for averaging, and several works applied Federated Learning to the recommender systems. Qi \textit{et al.}~\cite{qi2020privacy} applied Federated Learning and differential privacy to News Recommendation, which achieved a promising trade-off between recommendation performance and privacy protection. Lin \textit{et al.}~\cite{lin2020meta} introduced a MetaMF to account for the storage, energy and communication bandwidth in mobile environments, which distributed the gradient computation of Federated Learning across cloud and devices. Niu~\textit{et al.}~\cite{niu2020billion} followed a similar split in gradient computation and designed a tunable privacy to the local submodel.  Nevertheless, we argue that privacy protection is an exemplar usage of conceivable on-device learning. How to explore the collaboration between the cloud modeling and the device modeling for mutual benefit of two sides is also meaningful in practice~\cite{kulkarni2020survey}. One previous work~\cite{lu2019collaborative} that has the similar intuition differs from our method in the way to handle the sparsity issue and their method has not considered the frequent checkpoint loading issue.
\section{The proposed framework}
% In this section, we first introduce the preliminaries, and then introduce the two modules, \emph{MetaPatch} and \emph{MoMoDistill}, in DCCL.

\subsection{Preliminary}
Given a recommendation dataset $\left\{(x_n, y_n)\right\}_{n=1,\dots,N}$, we target to learn a mapping function $f:x_n\in \mathbb{R}^D\longrightarrow y_n\in\mathbb{R}$ on the cloud side. Here, $x_n$ is the input $D$-dimensional feature that concatenates all available candidate features and user context, $y_n$ is the user implicit feedback (click or not) to the corresponding candidate and $N$ is the sample number. On the device side, each device (indexed by $m$) has its own local dataset, $\left\{(x_n^{(m)}, y_n^{(m)})\right\}_{n=1,\dots, N^{(m)}}$. We add a few parameter-efficient patches~\cite{yuan2020parameter} to the cloud model $f$ (freezing its parameters on the 
device side) for each device to construct a new function $f^{(m)}:x_n^{(m)}\in \mathbb{R}^D\longrightarrow y_n^{(m)}\in\mathbb{R}$. Note that, only the parameters related to patches are used for the model personalization and updated based on the local dataset.

As summarized in previous sections, the works regarding \emph{on-device inference} and those related to \emph{on-device learning for the centralized model} have been explored in an independent perspective by means of the pervasive computing advantages. There are few works that consider how to make the device modeling and the cloud modeling to benefit both sides jointly in the recommendation scenarios. However, this is critical and meaningful, since the conventional centralized cloud model follows the ``model-over-data" paradigm and is prone to Matthew effect~\cite{perc2014matthew}. On the other side, personalized models updated with relatively limited samples on the device side  suffers from the local optimum~\cite{finn2017model}, which can be calibrated by the centralized cloud model. 
% \begin{comment}catastrophic forgetting~\cite{kirkpatrick2017overcoming} or\end{comment}
We introduce the DCCL to bridge the gap, which customizes the personalization on the device side and enhances the centralized model with the ``model-over-models" distillation over billions of personalized device models. Figure~\ref{fig:model} gives the general illustration of the DCCL framework, however, real-world challenges make it not straightforward for implementation. In the following, we will discuss the practical challenges and deployment of each side in DCCL.

\subsection{MetaPatch for On-device Personalization}
\begin{figure}[!t]
    \centering
    \includegraphics[width=0.45\textwidth]{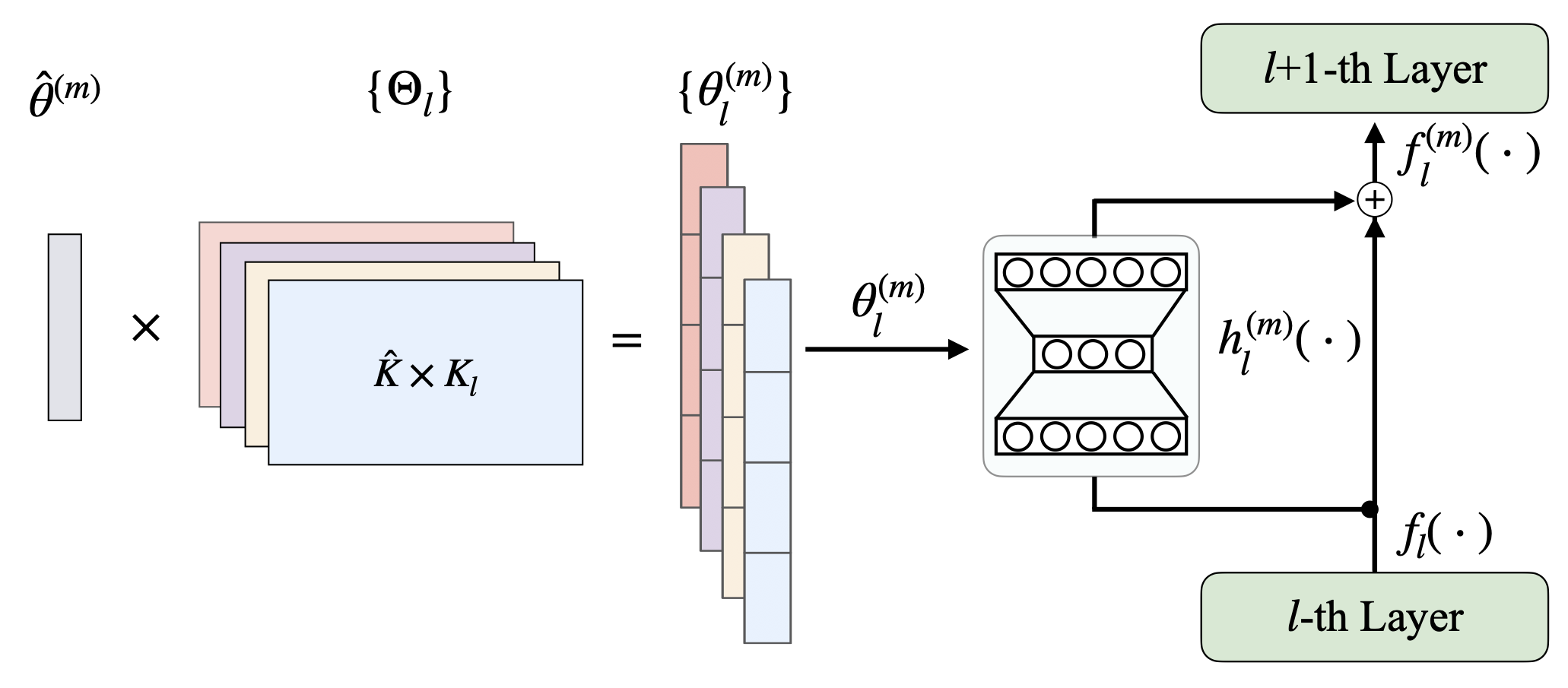}
    \caption{The proposed \emph{MetaPatch} to reduce the parameter space on the device side. It uses the global parameter basis $\Theta_l$ with the metapatch parameter $\hat{\theta}^{(m)}$ to generate all parameters $\{\theta_l^{(m)}\}$, and parameterize $h_l^{(m)}(\cdot)$ correspondingly. The dimension of patch parameters is reduced to $\hat{K}$ from $\sum_l K_l$.}
    \label{fig:metapatch}
\end{figure}
 Although the device hardware has been greatly improved in the recent years, compared to the rich computing power, energy and storage in the cloud, it is still resource-constrained to learn a complete big model on the device. Considering the parameter scale of the centralized cloud model $f$ and the storage of the intermediate activations, it is an impractical choice to adapt the whole network on the device side. Meanwhile, only finetuning last few layers is performance-limited due to the feature basis of the pretrained layers. Fortunately, some previous works have demonstrated that it is possible to achieve the comparable performance as the whole network finetuning via patch learning~\cite{cai2020tinytl,yuan2020parameter,houlsby2019parameter}. Inspired by these works, we insert the model patches on basis of the cloud model $f$ for on-device personalization. Formally, the output of the $l$-th layer attached with one patch on the $m$-th device is expressed as  
 \begin{align}\label{eq:patch}
    \begin{split}
        & f^{(m)}_l (\cdot) = f_l(\cdot) +  \underbrace{h^{(m)}_l}_{\text{patch}}(\cdot)\circ f_l(\cdot),
    \end{split}
 \end{align}
where LHS of Eq.\eqref{eq:patch} is the sum of the original $f_l(\cdot)$ and the patch response of $f_l(\cdot)$. Here, $h^{(m)}_l(\cdot)$ is the trainable patch function and $\circ$ denotes the function composition that treats the output of the previous function as the input. This construction facilitates that we can set a manual gate to mask the patch response and degenerate to the on-device inference with the centralized cloud model, anytime we want to ease the adaptation effect. Note that, the model patch could have different neural architectures. Here, we do not explore its variants but specify the same bottleneck architecture like~\cite{houlsby2019parameter}.

Nevertheless, we empirically find that the parameter space of multiple patches is still relatively too large to fit the sparse local samples. To overcome this issue, we propose a novel method \emph{MetaPatch} to reduce the parameter space. It is a kind of meta learning methods to generate parameters~\cite{ha2016hypernetworks,jia2016dynamic}, which shares the global parameter basis to reduce the parameters to be learned, thus more suits the on-device personalization. Concretely, assume the parameters of each patch are denoted by $\theta^{(m)}_l\in \mathbb{R}^{K_l}$ (flatten all parameters in the patch into a vector). Then, we can deduce the following decomposition
\begin{align} \label{eq:metapatch}
    \begin{split}
        & \theta^{(m)}_l = \Theta_l * \hat{\theta}^{(m)},
    \end{split}
\end{align}
where $\Theta_l \in \mathbb{R}^{K_l\times \hat{K}}$ is the globally shared parameter basis (freezing it on the device and learned in the cloud) and $\hat{\theta}^{(m)} \in \mathbb{R}^{\hat{K}}$ is the surrogate tunable parameter vector to generate each patch parameter $\theta^{(m)}_l$ in the device-model $f^{(m)}$. To facilitate the understanding, we term $\hat{\theta}^{(m)}$ as the metapatch parameter. In this paper, we keep $\sum_lK_l \gg \hat{K}$ so that the metapatch parameters to be learned for personalization are greatly reduced. Figure~\ref{fig:metapatch} illustrates the idea of \emph{MetaPatch}. Note that, regarding the pretraining of $\Theta_l$, we leave the discussion in the following section to avoid the clutter, since it is learned on the cloud side. According to Eq.~\eqref{eq:metapatch}, we implement the patch parameter generation via the metapatch parameter $\hat{\theta}^{(m)}$ instead of directly learning $\theta^{(m)}$. To learn the metapatch parameter, we can leverage the local dataset to minimize the following loss.
\begin{align} \label{eq:local}
    \begin{split}
        & \min_{\hat{\theta}^{(m)}} \ell{\left(y, \tilde{y}\right)}\big|_{\tilde{y}=f^{(m)}\left(x\right)},
    \end{split}
\end{align}
where $\ell$ is the pointwise cross-entropy loss, $f^{(m)}(\cdot)=f^{(m)}_L(\cdot)\circ\cdots f^{(m)}_l(\cdot)\cdots\circ f^{(m)}_1(\cdot)$ and $L$ is the number of total layers. After training the device specific parameter $\hat{\theta}^{(m)}$ by Eq.\eqref{eq:local}, we can use Eq.~\eqref{eq:metapatch} to generate all patches, and then insert them into the cloud network $f$ via Eq.\eqref{eq:patch} to get the final personalized model $f^{(m)}$, which will provide the on-device personalized recommendation. 

\subsection{MoMoDistill to Enhance the Cloud Modeling}
The conventional incremental training of the centralized cloud model follows the ``model-over-data'' paradigm. That is, when the new training samples are collected from devices, we directly perform the incremental learning based on the model trained in the early sample collection. The objective function is formulated as
\begin{align}\label{eq:it}
\begin{split}
    & \min_{\mathbf{W}_f} \ell\left(y, \hat{y}\right)\big|_{\hat{y}=f(x)},
\end{split}
\end{align}
where $\mathbf{W}_f$ is the network parameter of the cloud model $f$ to be trained. This is an independent perspective without considering the device modeling. However, the on-device personalization actually can be more powerful than the centralized cloud model to handle the corresponding local samples. Thus, the guidance from the on-device models could be a meaningful prior to help the cloud modeling. Inspired by this, we propose a ``model-over-models'' paradigm to simultaneously learn from data and aggregate the knowledge from on-device models, to enhance the training of the centralized cloud model. Besides, considering that on-device learning in a long term suffers from local optimization~\cite{finn2017model}, this step also helps calibrate the global model $f$ for the better on-device personalization. Formally, the objective with the distillation procedure on the samples from all devices is defined as, 
\begin{align}\label{eq:betadistill}
\begin{split}
    & \min_{\mathbf{W}_f} \ell\left(y, \hat{y}\right) + \beta~ \text{KL}(\tilde{y},\hat{y} )\big|_{\hat{y}=f(x),\tilde{y}=f^{(m)}(x)},
\end{split}
\end{align}
where $\beta$ is the hyperparameter to balance the distillation and ``model-over-data'' learning. 
Note that, the feasibility of the distillation in Eq.~\eqref{eq:betadistill} critically depends on the patch mechanism in the previous section, since it allows us to input the metapatch parameters like features with only loading the other model parameters of $f^{(m)}$ in one time. Otherwise, we will suffer from the engineering issue of reloading numerous checkpoints frequently, which is almost impossible for current open source frameworks like Tensorflow. 

\begin{figure}[!t]
    \centering
    \includegraphics[width=0.45\textwidth]{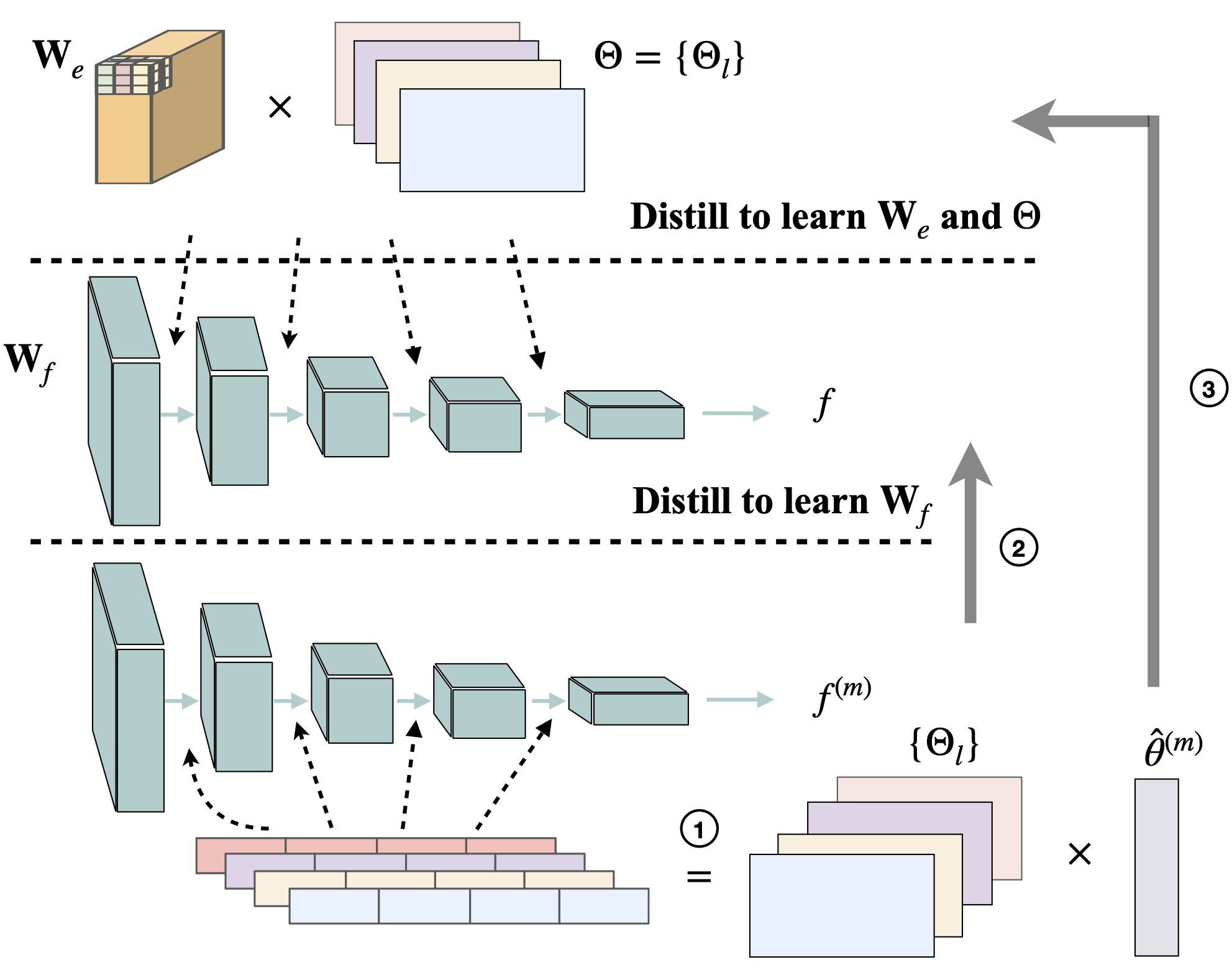}
    \caption{The proposed \emph{MoMoDistill} to enhance the cloud modeling. \textcircled{1} It leverages Eq.~\eqref{eq:metapatch} to compute the patch parameters and then constructs $\{f^{(m)}\}$. \textcircled{2} Following Eq.~\eqref{eq:betadistill}, we learn the centralized cloud model $f$ by model-over-models distillation from the on-device models $\{f^{(m)}\}$. \textcircled{3} Fixing $f$, optimize Eq.~\eqref{eq:betadistill2} to learn the global parameter basis $\Theta$.}
    \label{fig:betadistill}
\end{figure}

In \emph{MetaPatch}, we introduce the global parameter basis $\{\Theta_l\}$ (simplified by $\Theta$) to reduce the parameter space on the device. Regarding its training, we empirically find that coupled learning with $\bf{W}_f$ easily falls into undesirable local optimal, since they play different roles in terms of their semantics. Therefore, we resort to a progressive optimization strategy, that is, first optimize f based on Eq.~\eqref{eq:betadistill}, and then distill the knowledge for the parameter basis $\Theta$ with the learned $f$. For the second step, we design an auxiliary component by considering the heterogeneous characteristics of the metapatches from all devices and the cold-start issue at the beginning.
Concretely, given the dataset $\{(x,y,u^{(I(x))}, \hat{\theta}^{(I(x))})\}_{n=1,\dots,N}$, where $I$ maps the sample index to the device index and $u\subset x$ is the user profile features (\emph{e.g.}, age, gender, purchase level, etc) of the corresponding device, we define the following auxiliary encoder,
\begin{align} \label{eq:surrogate_updater}
    \begin{split}
        & U(\hat{\theta}, u)=\mathbf{W}^{(1)}\mathrm{tanh}(\mathbf{W}^{(2)}\hat{\theta} + \mathbf{W}^{(3)}u),
    \end{split}
\end{align}
where $\mathbf{W}^{(1)}\in \mathbb{R}^{\hat{K}_l\times\hat{K}_l}$, $\mathbf{W}^{(2)}\in \mathbb{R}^{\hat{K}_l\times \hat{K}_l }$, $\mathbf{W}^{(3)}\in \mathbb{R}^{\hat{K}_l\times d_u }$ are tunable projection matrices, $d_u$ is the dimension of user profile features. Here, we use $\bf{W}_e$ denotes the collection $\{\mathbf{W}^{(1)}, \mathbf{W}^{(2)}, \mathbf{W}^{(3)}\}$ for simplicity. To learn the global parameter basis, we replace $\hat{\theta}$ by $U(\hat{\theta}, u)$ to simulate Eq.~\eqref{eq:metapatch} to generate the model patch, \textit{i.e.}, $\Theta*U(\hat{\theta}, u)$, since actually $\hat{\theta}$ is too heterogeneous to be directly used. Then, combining $\Theta*U(\hat{\theta}, u)$ with $f$ learned in the first distillation step, we can form a new proxy device model $\hat{f}^{(m)}$ (different from $f^{(m)}$ in the patch generation). Here, we leverage such a proxy $\hat{f}^{(m)}$ to directly distill the knowledge from the true $f^{(m)}$ from devices, which optimizes $\Theta$ and $\bf{W}_e$ of the auxiliary encoder as follows,
\begin{align} \label{eq:betadistill2}
    \begin{split}
        & \min_{(\Theta,~\mathbf{W}_e
)} \ell{\left(y, \hat{y}\right)}+ \beta~ \text{KL}(\tilde{y},\hat{y} )\big|_{\hat{y}=\hat{f}^{(m)}(x),\tilde{y}=f^{(m)}(x)},
    \end{split}
\end{align}
Eq.~\eqref{eq:betadistill} and Eq.~\eqref{eq:betadistill2} progressively help learn the centralized cloud model and the global parameter basis. We specially term this progressive distillation mechanism as \emph{MoMoDistill} to emphasize our ``model-over-models'' paradigm different from the conventional ``model-over-data'' incremental training on the cloud side, and gives the illustration in Figure~\ref{fig:betadistill}. Finally, in Algorithm \ref{algo:1}, we summarize the complete procedure of DCCL with \emph{MetaPatch} and \emph{MoMoDistill}.

\begin{algorithm}[!t]
\SetAlgoLined
Pretrain the cloud model $f$, and then learn the global parameter basis $\Theta$ based on Eq.~\eqref{eq:betadistill2} by setting $\hat{\theta}$ as 0. \\
\While{lifecycle}{
\qquad Send $f$ and $\Theta$ to devices. \\
\qquad \textbf{Device}($f$, $\Theta$): \colorbox{gray!30}{$\rhd$~\emph{MetaPatch}} \\
\qquad\qquad 1) Accumulate the local data into batches \\
\qquad\qquad 2) On-device personalization via Eq.\eqref{eq:local} \\
\qquad\qquad 3) If time $>$ threshold: upload $f^{(m)}$ \\
\qquad\qquad 4) Else: return step 1) \\
\qquad Recycle all model patches $\{\hat{\theta}^{(m)}\}$. \\
\qquad\textbf{Cloud}($\{\hat{\theta}^{(m)}\}$): \colorbox{gray!30}{$\rhd$~\emph{MoMoDistill}} \\
\qquad\qquad 1) Optimize the cloud model $f$ based on Eq.\eqref{eq:betadistill} \\
\qquad\qquad 2) Learn the parameter basis $\Theta$ by Eq.\eqref{eq:betadistill2} \\
}
\caption{\mbox{Device-Cloud Collaborative Learning}}
\label{algo:1}
\end{algorithm}

\section{Experiments}
In this section, we will conduct a range of recommendation experiments to demonstrate the effectiveness of the proposed framework. To be specific, we will answer the following questions about DCCL.
\begin{enumerate}
    \item \textbf{RQ1:} Whether learning through the proposed framework can achieve a better performance compared with the conventional centralized incremental training in the cloud? The challenges come into two folds. First, we need to train the personalized models for each device. The detailed results of this part will be analysed in RQ2. Second, it is challenging to aggregate the knowledge from plenty of weakly heterogeneous on-device model ``experts", which 
    has never been explored before.
    
    \item \textbf{RQ2:} Whether on-device model personalization can achieve further improvement than the centralized cloud model? The sparse data flow on the local devices can be very challenging to learn and easy to overfit. To our best knowledge, this has never been investigated on the large-scale industrial recommendation datasets at device granularity. 

    \item \textbf{RQ3:} How is the convergence property regarding the their inter-loops between two modules and is it capable to maintain a cyclical process which iterates and optimizes between the cloud and the devices? It is important to characterize the long-term performance of our method, as once deployed, this framework will continually maintain such a learning cycle, which is a novel trial in the area of recommender systems.
\end{enumerate}

\subsection{Experimental Setup}
\subsubsection{Datasets}
Our experiments are implemented on three recommendation datasets Amazon, Movielens-1M and Taobao. The statistics of all the above datasets are shown in Table \ref{table:Statistics}. Generally, all these three datasets are user interactive histroy in sequence format , and the last user interacted item is cut out as test sample. For each last interacted item, we randomly sample 100 items that didn't appear in user's history. Detailed description about the data generation process is given in Appendix section.

\subsubsection{Baselines \& Our methods}
We compare DCCL with some classical cloud models in different perspectives, namely, the conventional methods, deep learning-based methods and sequence-based methods. 
\begin{itemize}
\item \textbf{Conventional methods:}
    \begin{itemize}
    \item \textbf{MF}~\cite{koren2009matrix}: The matrix factorization approach to model user-item interactions, which decomposes the observation as the product of the user embedding and the item embedding.
    \item \textbf{FM}~\cite{rendle2010factorization}: Factorization Machines (FM) maps real-valued features into a low-dimensional latent space, and models all interactions between variables using factorized parameters.
    \end{itemize}
\item \textbf{Deep learning-based methods:}
    \begin{itemize}
    \item \textbf{NeuMF}~\cite{he2017neural}: It generalizes the classifical MF into a deep-learning counterpart. It uses multiple neural layers to build a more expressive distance measure instead of inner product to capture the non-linearity in the implicit feedbacks.
    \item \textbf{DeepFM}~\cite{guo2017deepfm}: A wide \& deep architecture that incorporates the factorization machine to automatically extract the wide features instead of the previous hand-crafted effort.
    \end{itemize}
\item \textbf{Sequence-based methods:}
    \begin{itemize}
    \item \textbf{SASRec}~\cite{kang2018self}: A self-attention based sequential model, which utilizes the attention mechanism to adaptively assign weights to previous items for the next-item prediction.
    \item \textbf{DIN}~\cite{zhou2018deep}: A target-attention based user interest model, which takes the candidates as the query \textit{w.r.t.} the user historical behaviors to learn a user representation for prediction. 
    \end{itemize}
\end{itemize}

\begin{table}[!t]
\caption{Statistics of datasets used in this paper.}
\small
\centering
\begin{tabular}{l|c|c|c}
\toprule
    Dataset & Amazon & MovieLens-1M & Taobao \\ 
\midrule
Users & 360,828 & 6,040 & 17,018,570 \\ 
Goods & 68,000 & 3,900 & 20,000,000 \\
Categories & 257 & 21 & 14,641\\
Samples & 3,087,820 & 1,000,209 & $\sim$9 billions \\ 
\bottomrule
\end{tabular} \label{table:Statistics}
\end{table}

\begin{table*}[!t]
	\caption{The top-K recommendation performance of DCCL and baselines on three datasets.}
	\label{tab:performance}
	\centering
	{
	\begin{tabular}{l | l c c c c c c c c}
	\toprule
	Datasets & Metric & MF & FM & NeuMF & DeepFM & SASRec & DIN & DCCL & Improv. \\
	\midrule
	\multirow{5}{*}{Amazon}     & HitRate@1  & 23.69 & 21.53 & 26.10 & 25.43 & 26.53 & 26.56 & \textbf{26.94} & 1.43\% \\
								& HitRate@5  & 35.74 & 36.74 & 42.98 & 42.48 & 44.22 & 44.00 & \textbf{44.79} & 1.29\% \\
								& HitRate@10 & 44.38 & 47.90 & 52.32 & 53.51 & 54.94 & 55.43 & \textbf{56.59} & 2.09\% \\
								& NDCG@5     & 29.83 & 29.17 & 34.74 & 34.12 & 35.60 & 35.48 & \textbf{36.95} & 3.79\% \\
								& NDCG@10    & 32.61 & 32.77 & 37.76 & 37.67 & 39.07 & 39.22 & \textbf{40.45} & 3.14\% \\
	\midrule
    \multirow{5}{*}{MovieLens-1M}  & HitRate@1  & 14.60 & 14.90 & 16.45 & 15.41 & 34.85 & 37.45 & \textbf{38.69} & 3.31\% \\
								& HitRate@5  & 44.85 & 47.13 & 46.24 & 47.35 & 69.17 & 70.71 & \textbf{71.97} & 1.78\% \\
								& HitRate@10 & 63.54 & 64.40 & 65.36 & 65.46 & 80.69 & 81.25 & \textbf{82.23} & 1.21\% \\
								& NDCG@5     & 29.87 & 30.27 & 31.71 & 31.82 & 53.18 & 55.22 & \textbf{56.43} & 2.19\% \\
								& NDCG@10    & 35.89 & 36.49 & 37.90 & 37.68 & 56.94 & 58.65 & \textbf{59.77} & 1.91\% \\
	\midrule							
	\multirow{5}{*}{Taobao}		& HitRate@1  & 24.88 & 25.29 & 29.11 & 33.28 & 35.19 & 52.17 & \textbf{55.71} & 6.79\% \\
								& HitRate@5  & 50.83 & 51.18 & 55.42 & 57.26 & 60.13 & 68.12 & \textbf{70.31} & 3.21\% \\
								& HitRate@10 & 62.28 & 63.96 & 65.78 & 66.09 & 69.30 & 74.80 & \textbf{76.70} & 2.54\% \\
								& NDCG@5     & 38.46 & 38.80 & 43.03 & 46.09 & 48.52 & 60.65 & \textbf{63.42} & 4.57\% \\
								& NDCG@10    & 42.17 & 42.93 & 46.40 & 48.95 & 51.50 & 62.81 & \textbf{65.49} & 4.27\% \\
	\bottomrule
	\end{tabular}
	}
\end{table*}

For the whole experiments, we implement our model on the basis of \textbf{DIN}, where we insert the model patches in the last second fully-connected layer and the first two fully-connected layers after the feature embedding layer. In all comparisons, we term \emph{MetaPatch} as \textbf{DCCL-e}, and \emph{MoMoDistill} as \textbf{DCCL-m}, since the whole framework resembles EM\footnote{\url{https://en.wikipedia.org/wiki/Expectation-maximization_algorithm}} iterations. The default method to compare the baselines is named \textbf{DCCL}, which indicates that it goes through both on-device personalization and the ``model-over-models'' distillation.

\subsubsection{Evaluation Metrics}
The model performance in our experiments are measured by the widely used AUC
and NDCG
. They are respectively calculated by the following equations.
\begin{align} \label{auc}
\begin{split}
    & \text{HitRate}@K = \frac{1}{|\mathcal{U}|}\sum_{u\in \mathcal{U}} \mathds{1}(R_{u,g_u}\leq K),\\
    & \text{NDCG}@K = \sum_{u\in \mathcal{U}}  \frac 1 {|\mathcal{U}|}  \frac{2^{\mathds{1}(R_{u,g_u}\leq K)}-1}{\log_2(\mathds{1}(R_{u,g_u}\leq K)+1)},\\
    & \text{macro-AUC} = \sum_{u\in\mathcal{U}} \frac 1 {|\mathcal{U}|} \frac{\sum_{x_0\in D_T^{(u)}} \sum_{x_1 \in D_F^{(u)}}\mathbf{1}[f(x_1)<f(x_0)]}{|D_T^{(u)}||D_F^{(u)}|} ,\\
\end{split}
\end{align}
where $\mathcal{U}$ is the user set, $\mathds{1}(\cdot)$ is the indicator function, $R_{u,g_u}$ is the rank generated by the model for the ground truth item $g_u$ and user $u$, $f$ is the model to be evaluated and $D_T^{(u)}$, $D_F^{(u)}$ is the positive and negative sample sets in testing data.

\begin{figure*}[!t]
    \centering
    \includegraphics[width=0.92\textwidth]{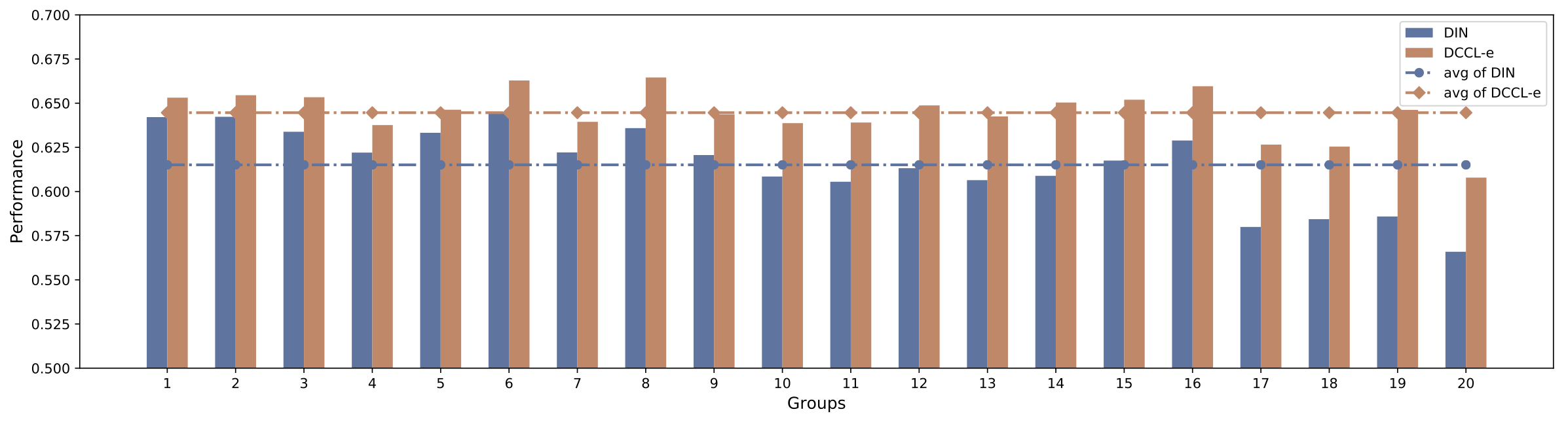}
    \caption{The macro-AUC of on-device personalization \textit{v.s.} DIN in all user groups. The performance of DIN in the head user groups are approximately better than that in the long-tailed user groups. In comparsion, DCCL-e performs better than DIN in all groups and specifically, achieves the larger improvement in the long-tailed user groups.}
    \label{fig:rq1}
\end{figure*}

\subsection{Experiments and Results}

\subsubsection{RQ1}
To demonstrate the effectiveness of DCCL, we conduct the experiments on Amazon, Movielens and Taobao to compare to a range of baselines. Aligned with the popular experimental settings~\cite{he2017neural,zhou2018deep}, the last interactive item of each user on three datasets is left for evaluation and all items before the last one are used for training. For DCCL, we split the training data into two parts on average according to the temporal order: one part is for the pretraining of the backbone (DIN) and the other part is for the training of DCCL. In the experiments, we conduct one-round DCCL-e and DCCL-m. Finally, the DCCL-m is compared with the six representative models, whose results are summarized in Table~\ref{tab:performance}.

According to Table~\ref{tab:performance}, we find that the deep learning based methods NeuMF and DeepFM usually outperform the conventional methods MF and FM, and the sequence-based methods SASRec and DIN consistently outperform previous non-sequence-based methods. Our DCCL builds upon on the best baseline DIN and further improves its results. Specifically, DCCL shows about $2\%$ or more improvements in terms of NDCG@10, and at least $1\%$ improvements in terms of HitRate@10 on all three datasets. The performance on both small and large datasets confirms the superiority of DCCL.

\subsubsection{RQ2}
In this section, we target to demonstrate that how on-device personalization via \emph{MetaPatch} (abbreviated as DCCl-e) can improve the recommendation performance from different levels of users compared with the centralized cloud model. Considering the data scale and the availability of the context information for visualization, only the Taobao dataset is used to conduct this experiment. To validate the performance of DCCL-e in the fine-grained granularity, we sort the users based on their sample numbers and then partition them into 20 groups on average along the sorted user axis (see the statistic of the sample number \textit{w.r.t.} the user in the appendix). After on-device model personalization, we calculate the performance for each group based on the personalized models. Here, the macro-AUC is used, which equally treats the users in the group instead of the sample-number-aware group AUC~\cite{zhou2018deep}. 

We use DIN as baseline in this experiment and pretrain it on the Taobao Dataset of the first 20 days. Then, we test the model in the data of the remaining 10 days. For DCCL-e, we first pretrain DIN on the Taobao Dataset of the first 10 days, and then insert the patches into the pretrained DIN same as previous settings. Finally, we perform the on-device personalization in the subsequent 10 days. Similarly, we test the DCCL-e on the data of the last 10 days. The evaluation is respectively conducted in the 20 groups and their results are plot in Figure~\ref{fig:rq1}. According to the results, we can find that with the increase of the group index number, the performance approximately decreases. This is because the users in the group of larger indices are more like the long-tailed users based on our group partition, and their patterns are easily ignored or even sacrificed by the centralized cloud model. In comparison, DCCL-e shows the consistent improvement over DIN on all groups, and especially achieve the large improvement in the long-tailed user groups. 

\begin{figure}[!t]
    \centering
    \includegraphics[width=0.4\textwidth]{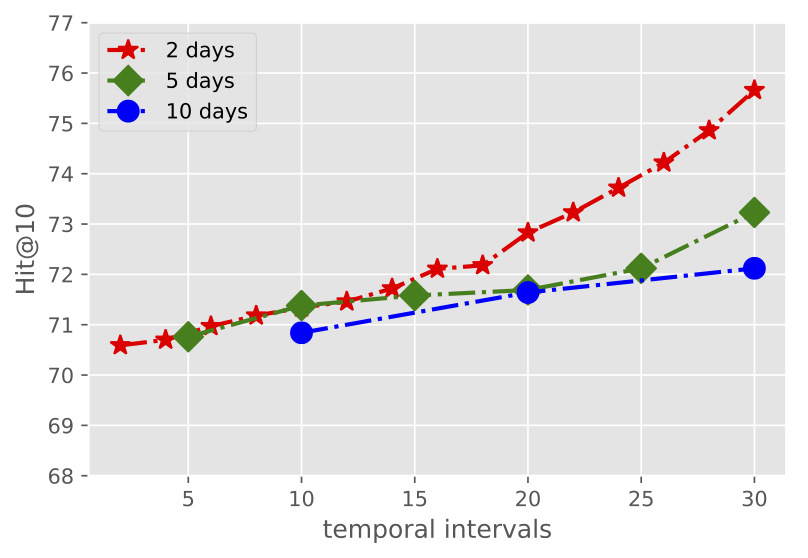}
    \caption{The convergence property of DCCL in three different temporal intervals of each round on the Taobao dataset.}
    \label{fig:rq3}
\end{figure}

\subsubsection{RQ3}
To illustrate the convergence property of DCCL, we conduct the experiments on the Taobao dataset in different device-cloud interaction temporal intervals. Concretely, we specify every 2, 5, 10 days interactions between device and cloud, and respectively trace the performance of each round evaluated on the last click of each user. Figure~\ref{fig:rq3} illustrates the convergence procedure of DCCL in different intervals. According to Figure~\ref{fig:rq3}, we observe that frequent interactions achieve much better performance than the infrequent counterparts. We speculate that, as \emph{MeatPatch} and \emph{MoMoDistill} could promote each other at every round, the advantages in performance have been continuously strengthened with more frequent interactions. However, the side effect is we have to frequently update the on-device models, which may introduce other uncertain crash risks. Thus, in the real-world scenarios, we still need to make a trade-off between performance and the interaction interval in the cold-start.

\begin{figure*}[!t]
    \centering
    \includegraphics[width=0.87\textwidth]{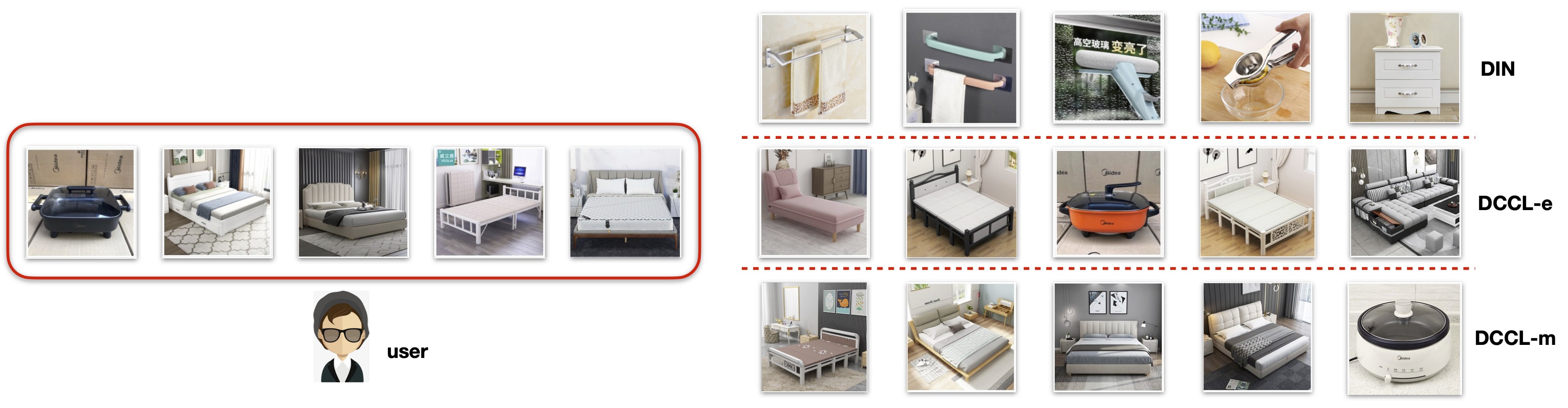}
    \caption{The case study to compare the recommendation of DIN, DCCL-e and DCCL-m for one long-tailed user. The left hand side presents five items that user 1 clicked and the right hand side gives the top-5 recommendation from three methods.}
    \label{fig:case_study}
\end{figure*}

\subsubsection{Ablation and Case Study}
In this section, we present more analysis of DCCL about \emph{MoMoDistill} and \emph{MetaPatch}. Then, we will exhibit a case study to illustrate the difference of recommendation results between DCCL and DIN for the long-tailed users.

\begin{table}[!h]
	\caption{The results of one-round DCCL compared to DIN.}
	\label{tab:e_m_compare}
	\centering
		{
	\begin{tabular}{l|*{3}{c} }
	\toprule
	\multicolumn{1}{l|}{Method} & 
	\multicolumn{1}{c}{DIN} &
	\multicolumn{1}{c}{DCCL-e} &
	\multicolumn{1}{c}{DCCL-m} 
	\\\midrule
	HitRate@1 & 52.17 & 55.03 & \bf{55.71} \\
    HitRate@5 & 68.12 & 70.03 & \bf{70.31} \\
    HitRate@10 & 74.80 & 76.46 & \bf{76.70} \\
    NDCG@5 & 60.65 & 62.99 & \bf{63.42} \\
    NDCG@10 & 62.81 & 65.07 & \bf{65.49} \\
	\bottomrule
	\end{tabular}
	}
\end{table}

For the first  study, we give the results of \emph{MetaPatch} and \emph{MoMoDistill} in one-round DCCL on the Taobao dataset and compare with DIN in Table~\ref{tab:e_m_compare}. From the results, we can observe the progressive improvement after DCCL-e and DCCL-m, and DCCL-e acquires more benefit than DCCL-m in terms of the improvement. The revenue behind DCCL-e is \emph{MetaPatch} customizes a peronsalized model for each user to improve their recommendation experience once new behavior logs are collected on device, without the delayed update from the centralized cloud server. Regarding DCCL-m, as stated before, the local samples could be scarce and noisy. Therefore, the lengthy on-device personalization will be easy to be trapped by the limited knowledge of each user and result in the local optimal for the recommendation model. The further improvements from DCCL-m confirm the necessity of \emph{MoMoDistill} to re-calibarate the backone and the parameter basis. However, if we conduct the experiments without our two modules, the model performance is as DIN, which is  not better than both DCCL-e and DCCL-m.

\begin{table}[!t]
	\caption{One-round DCCL with different positional patches.}
	\label{tab:patch_compare}
	\centering
		{
	\begin{tabular}{l|*{3}{c} }
	\toprule
	\multicolumn{1}{l|}{Position} & 
	\multicolumn{1}{c}{1st Junction} &
	\multicolumn{1}{c}{2nd Junction} &
	\multicolumn{1}{c}{3rd Junction} 
	\\\midrule
    HitRate@1  & 53.26 & \bf{54.10}  & 52.36 \\
    HitRate@5  & 68.89 & \bf{69.36}  & 68.14 \\
    HitRate@10 & 75.54 & \bf{75.86}  & 74.85 \\
    NDCG@5     & 61.56 & \bf{62.19}  & 60.74 \\
    NDCG@10    & 63.71 & \bf{64.29}  & 62.92 \\
	\bottomrule
	\end{tabular}
	}
\end{table}

For the second ablation study, we explore the effect of the model patches in different layer junctions. As claimed in previous sections, we insert two patches (1st Junction, 2nd Junction) in the two fully-connected layers respectively after the feature embedding layer, and one patch (3rd Junction) to the layer before the last softmax transformation layer. In this experiment, we validate their effectiveness by only keep each of them in one-round DCCL. Their results on the Taobao dataset are summarized in Table~\ref{tab:patch_compare}. Compared with the full model in Table~\ref{tab:e_m_compare}, we can find that removing the model patch would decrease the performance. And the results suggest that the model patches in the 1st and 2nd juncitons are more effective than the higher layer (the 3rd junction). The intuition is that the model patches in low layers 
more easily adjust the item contributions, which highlights or downweights items according to the user interests, resulting in a more expressive personalization.

\begin{figure}[!t]
    \centering
    \includegraphics[width=0.4\textwidth]{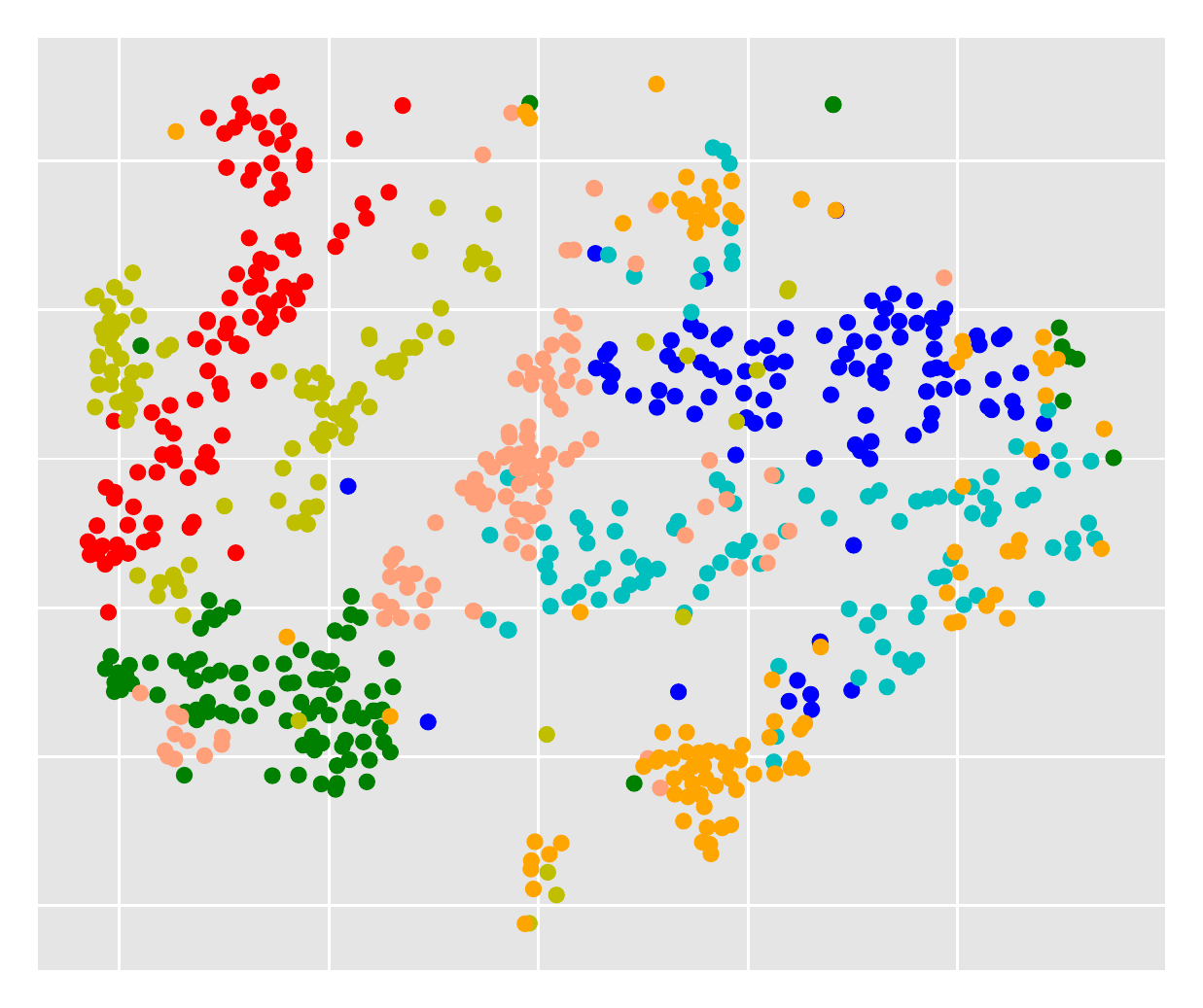}
    \caption{T-SNE visualization of the metapatches. The color is the representative Level-1 category to train metapatches.}
    \label{fig:tsne}
\end{figure}

To visualize the metapatches learned from on-device personalization, we project their parameter vectors $\{\hat{\theta}^{(m)}\}$ into the 2-D space via t-SNE\footnote{\url{https://en.wikipedia.org/wiki/T-distributed_stochastic_neighbor_embedding}} as shown in Figure~\ref{fig:tsne}. For a better visualization, we assign each point with the color representing the most frequent level-1 category in the local samples to train the metapatch. Note that, it does not mean the true label but we conjecture that the optimization direction of the metapatch is possibly correlated with the sample categories. From Figure~\ref{fig:tsne}, we can find that the points belonging to the same color approximately form into a cluster. This might indicate they share similar on-device models to some extent, which inherently keeps the consistency with motivation of learning the global parameter basis $\Theta$ to reduce the parameter space.

To visualize the difference between the centralized cloud model and DCCL, we illustrate one case study based on DIN, DCCL-e and DCCL-m in Figure~\ref{fig:case_study}. Specifically, one long-tailed user is chosen to validate their effectiveness. As shown in Figure~\ref{fig:case_study}, although the user has clicked many items from the bed categories, DIN has not recommend the expected items related to his history. It is because the bed category is a minority compared to other daily supplies, and the centralized cloud model are optimized bias to this pattern, which is prone to recall other supplies. In comparison, DCCL-e that has adapted to the historical behaviors of this user recommends more relevant items of this category. Nevertheless, it is not perfect to ease the bias effect of the pretrained cloud model and also introduces some bad cases like the last sofa item. In comparison, DCCL-m that refines the global parameter basis further alleviates this issue and recommends all items to the user historical behaviors. 
\section{Conclusion}
In this paper, we focus on the general mobile-cloud collaboration scenarios, and propose a Device-Cloud Collaborative Learning framework to explore mutual benefit of the mobile and the cloud modeling. Specifically, we introduce a novel \emph{MetaPatch} approach to efficiently achieve the model personalization for each device, and simultaneously introduce a \emph{MoMoDistill} enhance the cloud modeling via the knowledge distillation from the personalized on-device models. Extensive experiments on a range of datasets demonstrate the superiority of DCCL over state-of-the-art methods. However, as an initial exploration about the collaboration between the device and the cloud modeling, more works need to be contributed to on-device intelligence and solving the general challenges about the data sparsity, noise and other computational constraints.

\section{ACKNOWLEDGMENTS}
This work is supported by National Natural Science Foundation of China under Grant U20A20222.

\bibliographystyle{ACM-Reference-Format}
\bibliography{acmart}


\begin{thebibliography}{49}

%%% the Web address.

\ifx \showCODEN    \undefined \def \showCODEN     #1{\unskip}     \fi
\ifx \showDOI      \undefined \def \showDOI       #1{#1}\fi
\ifx \showISBNx    \undefined \def \showISBNx     #1{\unskip}     \fi
\ifx \showISBNxiii \undefined \def \showISBNxiii  #1{\unskip}     \fi
\ifx \showISSN     \undefined \def \showISSN      #1{\unskip}     \fi
\ifx \showLCCN     \undefined \def \showLCCN      #1{\unskip}     \fi
\ifx \shownote     \undefined \def \shownote      #1{#1}          \fi
\ifx \showarticletitle \undefined \def \showarticletitle #1{#1}   \fi
\ifx \showURL      \undefined \def \showURL       {\relax}        \fi
% invisible to TeX
\providecommand\bibfield[2]{#2}
\providecommand\bibinfo[2]{#2}
\providecommand\natexlab[1]{#1}
\providecommand\showeprint[2][]{arXiv:#2}

\bibitem[\protect\citeauthoryear{Bedi, Venayagamoorthy, Singh, Brooks, and
  Wang}{Bedi et~al\mbox{.}}{2018}]%
        {bedi2018review}
\bibfield{author}{\bibinfo{person}{Guneet Bedi}, \bibinfo{person}{Ganesh~Kumar
  Venayagamoorthy}, \bibinfo{person}{Rajendra Singh},
  \bibinfo{person}{Richard~R Brooks}, {and} \bibinfo{person}{Kuang-Ching
  Wang}.} \bibinfo{year}{2018}\natexlab{}.
\newblock \showarticletitle{Review of Internet of Things (IoT) in electric
  power and energy systems}.
\newblock \bibinfo{journal}{\emph{IEEE Internet of Things Journal}}
  \bibinfo{volume}{5}, \bibinfo{number}{2} (\bibinfo{year}{2018}),
  \bibinfo{pages}{847--870}.
\newblock


\bibitem[\protect\citeauthoryear{Bistritz, Mann, and Bambos}{Bistritz
  et~al\mbox{.}}{2020}]%
        {bistritz2020distributed}
\bibfield{author}{\bibinfo{person}{Ilai Bistritz}, \bibinfo{person}{Ariana
  Mann}, {and} \bibinfo{person}{Nicholas Bambos}.}
  \bibinfo{year}{2020}\natexlab{}.
\newblock \showarticletitle{Distributed Distillation for On-Device Learning}.
\newblock \bibinfo{journal}{\emph{Advances in Neural Information Processing
  Systems}}  \bibinfo{volume}{33} (\bibinfo{year}{2020}).
\newblock


\bibitem[\protect\citeauthoryear{Cai, Gan, Wang, Zhang, and Han}{Cai
  et~al\mbox{.}}{2020a}]%
        {cai2020once}
\bibfield{author}{\bibinfo{person}{Han Cai}, \bibinfo{person}{Chuang Gan},
  \bibinfo{person}{Tianzhe Wang}, \bibinfo{person}{Zhekai Zhang}, {and}
  \bibinfo{person}{Song Han}.} \bibinfo{year}{2020}\natexlab{a}.
\newblock \showarticletitle{Once for All: Train One Network and Specialize it
  for Efficient Deployment}. In \bibinfo{booktitle}{\emph{ICLR}}.
\newblock


\bibitem[\protect\citeauthoryear{Cai, Gan, Zhu, and Han}{Cai
  et~al\mbox{.}}{2020b}]%
        {cai2020tinytl}
\bibfield{author}{\bibinfo{person}{Han Cai}, \bibinfo{person}{Chuang Gan},
  \bibinfo{person}{Ligeng Zhu}, {and} \bibinfo{person}{Song Han}.}
  \bibinfo{year}{2020}\natexlab{b}.
\newblock \showarticletitle{TinyTL: Reduce Memory, Not Parameters for Efficient
  On-Device Learning}.
\newblock \bibinfo{journal}{\emph{Advances in Neural Information Processing
  Systems}}  \bibinfo{volume}{33} (\bibinfo{year}{2020}).
\newblock


\bibitem[\protect\citeauthoryear{Chen, Zhang, Tsang, Pan, and Su}{Chen
  et~al\mbox{.}}{2020}]%
        {chen2020towards}
\bibfield{author}{\bibinfo{person}{Xu Chen}, \bibinfo{person}{Ya Zhang},
  \bibinfo{person}{Ivor Tsang}, \bibinfo{person}{Yuangang Pan}, {and}
  \bibinfo{person}{Jingchao Su}.} \bibinfo{year}{2020}\natexlab{}.
\newblock \showarticletitle{Towards Equivalent Transformation of User
  Preferences in Cross Domain Recommendation}.
\newblock \bibinfo{journal}{\emph{arXiv preprint arXiv:2009.06884}}
  (\bibinfo{year}{2020}).
\newblock


\bibitem[\protect\citeauthoryear{Cheng, Koc, Harmsen, Shaked, Chandra, Aradhye,
  Anderson, Corrado, Chai, Ispir, et~al\mbox{.}}{Cheng et~al\mbox{.}}{2016}]%
        {cheng2016wide}
\bibfield{author}{\bibinfo{person}{Heng-Tze Cheng}, \bibinfo{person}{Levent
  Koc}, \bibinfo{person}{Jeremiah Harmsen}, \bibinfo{person}{Tal Shaked},
  \bibinfo{person}{Tushar Chandra}, \bibinfo{person}{Hrishi Aradhye},
  \bibinfo{person}{Glen Anderson}, \bibinfo{person}{Greg Corrado},
  \bibinfo{person}{Wei Chai}, \bibinfo{person}{Mustafa Ispir}, {et~al\mbox{.}}}
  \bibinfo{year}{2016}\natexlab{}.
\newblock \showarticletitle{Wide \& deep learning for recommender systems}. In
  \bibinfo{booktitle}{\emph{Proceedings of the 1st workshop on deep learning
  for recommender systems}}. \bibinfo{pages}{7--10}.
\newblock


\bibitem[\protect\citeauthoryear{Cui, Chen, Yao, and Zhang}{Cui
  et~al\mbox{.}}{2018}]%
        {cui2018variational}
\bibfield{author}{\bibinfo{person}{Kenan Cui}, \bibinfo{person}{Xu Chen},
  \bibinfo{person}{Jiangchao Yao}, {and} \bibinfo{person}{Ya Zhang}.}
  \bibinfo{year}{2018}\natexlab{}.
\newblock \showarticletitle{Variational collaborative learning for user
  probabilistic representation}.
\newblock \bibinfo{journal}{\emph{Workshop of the AAAI conference on artificial
  intelligence}} (\bibinfo{year}{2018}).
\newblock


\bibitem[\protect\citeauthoryear{Dai, Spasi{\'c}, Meyer, Chapman, and
  Andres}{Dai et~al\mbox{.}}{2019}]%
        {dai2019machine}
\bibfield{author}{\bibinfo{person}{Xiangfeng Dai}, \bibinfo{person}{Irena
  Spasi{\'c}}, \bibinfo{person}{Bradley Meyer}, \bibinfo{person}{Samuel
  Chapman}, {and} \bibinfo{person}{Frederic Andres}.}
  \bibinfo{year}{2019}\natexlab{}.
\newblock \showarticletitle{Machine learning on mobile: An on-device inference
  app for skin cancer detection}. In \bibinfo{booktitle}{\emph{2019 Fourth
  International Conference on Fog and Mobile Edge Computing (FMEC)}}. IEEE,
  \bibinfo{pages}{301--305}.
\newblock


\bibitem[\protect\citeauthoryear{Eshratifar, Abrishami, and Pedram}{Eshratifar
  et~al\mbox{.}}{2019}]%
        {eshratifar2019jointdnn}
\bibfield{author}{\bibinfo{person}{Amir~Erfan Eshratifar},
  \bibinfo{person}{Mohammad~Saeed Abrishami}, {and} \bibinfo{person}{Massoud
  Pedram}.} \bibinfo{year}{2019}\natexlab{}.
\newblock \showarticletitle{JointDNN: an efficient training and inference
  engine for intelligent mobile cloud computing services}.
\newblock \bibinfo{journal}{\emph{IEEE Transactions on Mobile Computing}}
  (\bibinfo{year}{2019}).
\newblock


\bibitem[\protect\citeauthoryear{Finn, Abbeel, and Levine}{Finn
  et~al\mbox{.}}{2017}]%
        {finn2017model}
\bibfield{author}{\bibinfo{person}{Chelsea Finn}, \bibinfo{person}{Pieter
  Abbeel}, {and} \bibinfo{person}{Sergey Levine}.}
  \bibinfo{year}{2017}\natexlab{}.
\newblock \showarticletitle{Model-agnostic meta-learning for fast adaptation of
  deep networks}.
\newblock \bibinfo{journal}{\emph{ICML}} (\bibinfo{year}{2017}).
\newblock


\bibitem[\protect\citeauthoryear{Gong, Jiang, Feng, Hu, Zhao, Liu, and Ou}{Gong
  et~al\mbox{.}}{2020}]%
        {gong2020edgerec}
\bibfield{author}{\bibinfo{person}{Yu Gong}, \bibinfo{person}{Ziwen Jiang},
  \bibinfo{person}{Yufei Feng}, \bibinfo{person}{Binbin Hu},
  \bibinfo{person}{Kaiqi Zhao}, \bibinfo{person}{Qingwen Liu}, {and}
  \bibinfo{person}{Wenwu Ou}.} \bibinfo{year}{2020}\natexlab{}.
\newblock \showarticletitle{EdgeRec: Recommender System on Edge in Mobile
  Taobao}. In \bibinfo{booktitle}{\emph{Proceedings of the 29th ACM
  International Conference on Information \& Knowledge Management}}.
  \bibinfo{pages}{2477--2484}.
\newblock


\bibitem[\protect\citeauthoryear{Guo, Tang, Ye, Li, and He}{Guo
  et~al\mbox{.}}{2017}]%
        {guo2017deepfm}
\bibfield{author}{\bibinfo{person}{Huifeng Guo}, \bibinfo{person}{Ruiming
  Tang}, \bibinfo{person}{Yunming Ye}, \bibinfo{person}{Zhenguo Li}, {and}
  \bibinfo{person}{Xiuqiang He}.} \bibinfo{year}{2017}\natexlab{}.
\newblock \showarticletitle{DeepFM: a factorization-machine based neural
  network for CTR prediction}. In \bibinfo{booktitle}{\emph{Proceedings of the
  26th International Joint Conference on Artificial Intelligence}}.
  \bibinfo{pages}{1725--1731}.
\newblock


\bibitem[\protect\citeauthoryear{Ha, Dai, and Le}{Ha et~al\mbox{.}}{2017}]%
        {ha2016hypernetworks}
\bibfield{author}{\bibinfo{person}{David Ha}, \bibinfo{person}{Andrew Dai},
  {and} \bibinfo{person}{Quoc~V Le}.} \bibinfo{year}{2017}\natexlab{}.
\newblock \showarticletitle{Hypernetworks}.
\newblock \bibinfo{journal}{\emph{ICLR}} (\bibinfo{year}{2017}).
\newblock


\bibitem[\protect\citeauthoryear{Han, Mao, and Dally}{Han
  et~al\mbox{.}}{2016}]%
        {han2016deep}
\bibfield{author}{\bibinfo{person}{Song Han}, \bibinfo{person}{Huizi Mao},
  {and} \bibinfo{person}{William~J Dally}.} \bibinfo{year}{2016}\natexlab{}.
\newblock \showarticletitle{Deep compression: Compressing deep neural networks
  with pruning, trained quantization and huffman coding}.
\newblock \bibinfo{journal}{\emph{ICLR}} (\bibinfo{year}{2016}).
\newblock


\bibitem[\protect\citeauthoryear{Han, Pool, Tran, and Dally}{Han
  et~al\mbox{.}}{2015}]%
        {han2015learning}
\bibfield{author}{\bibinfo{person}{Song Han}, \bibinfo{person}{Jeff Pool},
  \bibinfo{person}{John Tran}, {and} \bibinfo{person}{William Dally}.}
  \bibinfo{year}{2015}\natexlab{}.
\newblock \showarticletitle{Learning both weights and connections for efficient
  neural network}.
\newblock \bibinfo{journal}{\emph{Advances in neural information processing
  systems}}  \bibinfo{volume}{28} (\bibinfo{year}{2015}),
  \bibinfo{pages}{1135--1143}.
\newblock


\bibitem[\protect\citeauthoryear{He, Liao, Zhang, Nie, Hu, and Chua}{He
  et~al\mbox{.}}{2017}]%
        {he2017neural}
\bibfield{author}{\bibinfo{person}{Xiangnan He}, \bibinfo{person}{Lizi Liao},
  \bibinfo{person}{Hanwang Zhang}, \bibinfo{person}{Liqiang Nie},
  \bibinfo{person}{Xia Hu}, {and} \bibinfo{person}{Tat-Seng Chua}.}
  \bibinfo{year}{2017}\natexlab{}.
\newblock \showarticletitle{Neural collaborative filtering}. In
  \bibinfo{booktitle}{\emph{Proceedings of the 26th international conference on
  world wide web}}. \bibinfo{pages}{173--182}.
\newblock


\bibitem[\protect\citeauthoryear{Houlsby, Giurgiu, Jastrzebski, Morrone,
  De~Laroussilhe, Gesmundo, Attariyan, and Gelly}{Houlsby
  et~al\mbox{.}}{2019}]%
        {houlsby2019parameter}
\bibfield{author}{\bibinfo{person}{Neil Houlsby}, \bibinfo{person}{Andrei
  Giurgiu}, \bibinfo{person}{Stanislaw Jastrzebski}, \bibinfo{person}{Bruna
  Morrone}, \bibinfo{person}{Quentin De~Laroussilhe}, \bibinfo{person}{Andrea
  Gesmundo}, \bibinfo{person}{Mona Attariyan}, {and} \bibinfo{person}{Sylvain
  Gelly}.} \bibinfo{year}{2019}\natexlab{}.
\newblock \showarticletitle{Parameter-efficient transfer learning for NLP}. In
  \bibinfo{booktitle}{\emph{International Conference on Machine Learning}}.
  PMLR, \bibinfo{pages}{2790--2799}.
\newblock


\bibitem[\protect\citeauthoryear{Howard, Zhu, Chen, Kalenichenko, Wang, Weyand,
  Andreetto, and Adam}{Howard et~al\mbox{.}}{2017}]%
        {howard2017mobilenets}
\bibfield{author}{\bibinfo{person}{Andrew~G Howard}, \bibinfo{person}{Menglong
  Zhu}, \bibinfo{person}{Bo Chen}, \bibinfo{person}{Dmitry Kalenichenko},
  \bibinfo{person}{Weijun Wang}, \bibinfo{person}{Tobias Weyand},
  \bibinfo{person}{Marco Andreetto}, {and} \bibinfo{person}{Hartwig Adam}.}
  \bibinfo{year}{2017}\natexlab{}.
\newblock \showarticletitle{Mobilenets: Efficient convolutional neural networks
  for mobile vision applications}.
\newblock \bibinfo{journal}{\emph{ICLR}} (\bibinfo{year}{2017}).
\newblock


\bibitem[\protect\citeauthoryear{Jannach and Ludewig}{Jannach and
  Ludewig}{2017}]%
        {jannach2017recurrent}
\bibfield{author}{\bibinfo{person}{Dietmar Jannach} {and}
  \bibinfo{person}{Malte Ludewig}.} \bibinfo{year}{2017}\natexlab{}.
\newblock \showarticletitle{When recurrent neural networks meet the
  neighborhood for session-based recommendation}. In
  \bibinfo{booktitle}{\emph{Proceedings of the Eleventh ACM Conference on
  Recommender Systems}}. \bibinfo{pages}{306--310}.
\newblock


\bibitem[\protect\citeauthoryear{Jia, De~Brabandere, Tuytelaars, and Gool}{Jia
  et~al\mbox{.}}{2016}]%
        {jia2016dynamic}
\bibfield{author}{\bibinfo{person}{Xu Jia}, \bibinfo{person}{Bert
  De~Brabandere}, \bibinfo{person}{Tinne Tuytelaars}, {and}
  \bibinfo{person}{Luc~V Gool}.} \bibinfo{year}{2016}\natexlab{}.
\newblock \showarticletitle{Dynamic filter networks}.
\newblock \bibinfo{journal}{\emph{Advances in neural information processing
  systems}}  \bibinfo{volume}{29} (\bibinfo{year}{2016}),
  \bibinfo{pages}{667--675}.
\newblock


\bibitem[\protect\citeauthoryear{Kang and McAuley}{Kang and McAuley}{2018}]%
        {kang2018self}
\bibfield{author}{\bibinfo{person}{Wang-Cheng Kang} {and}
  \bibinfo{person}{Julian McAuley}.} \bibinfo{year}{2018}\natexlab{}.
\newblock \showarticletitle{Self-attentive sequential recommendation}. In
  \bibinfo{booktitle}{\emph{2018 IEEE International Conference on Data Mining
  (ICDM)}}. IEEE, \bibinfo{pages}{197--206}.
\newblock


\bibitem[\protect\citeauthoryear{Karimireddy, Kale, Mohri, Reddi, Stich, and
  Suresh}{Karimireddy et~al\mbox{.}}{2020}]%
        {karimireddy2019scaffold}
\bibfield{author}{\bibinfo{person}{Sai~Praneeth Karimireddy},
  \bibinfo{person}{Satyen Kale}, \bibinfo{person}{Mehryar Mohri},
  \bibinfo{person}{Sashank~J Reddi}, \bibinfo{person}{Sebastian~U Stich}, {and}
  \bibinfo{person}{Ananda~Theertha Suresh}.} \bibinfo{year}{2020}\natexlab{}.
\newblock \showarticletitle{Scaffold: Stochastic controlled averaging for
  on-device federated learning}.
\newblock \bibinfo{journal}{\emph{ICML}} (\bibinfo{year}{2020}).
\newblock


\bibitem[\protect\citeauthoryear{Konečný, McMahan, Yu, Richtarik, Suresh, and
  Bacon}{Konečný et~al\mbox{.}}{2016}]%
        {45648federated}
\bibfield{author}{\bibinfo{person}{Jakub Konečný},
  \bibinfo{person}{H.~Brendan McMahan}, \bibinfo{person}{Felix~X. Yu},
  \bibinfo{person}{Peter Richtarik}, \bibinfo{person}{Ananda~Theertha Suresh},
  {and} \bibinfo{person}{Dave Bacon}.} \bibinfo{year}{2016}\natexlab{}.
\newblock \showarticletitle{Federated Learning: Strategies for Improving
  Communication Efficiency}. In \bibinfo{booktitle}{\emph{ICLR}}.
\newblock


\bibitem[\protect\citeauthoryear{Koren, Bell, and Volinsky}{Koren
  et~al\mbox{.}}{2009}]%
        {koren2009matrix}
\bibfield{author}{\bibinfo{person}{Yehuda Koren}, \bibinfo{person}{Robert
  Bell}, {and} \bibinfo{person}{Chris Volinsky}.}
  \bibinfo{year}{2009}\natexlab{}.
\newblock \showarticletitle{Matrix factorization techniques for recommender
  systems}.
\newblock \bibinfo{journal}{\emph{Computer}} \bibinfo{volume}{42},
  \bibinfo{number}{8} (\bibinfo{year}{2009}), \bibinfo{pages}{30--37}.
\newblock


\bibitem[\protect\citeauthoryear{Kulkarni, Kulkarni, and Pant}{Kulkarni
  et~al\mbox{.}}{2020}]%
        {kulkarni2020survey}
\bibfield{author}{\bibinfo{person}{Viraj Kulkarni}, \bibinfo{person}{Milind
  Kulkarni}, {and} \bibinfo{person}{Aniruddha Pant}.}
  \bibinfo{year}{2020}\natexlab{}.
\newblock \showarticletitle{Survey of Personalization Techniques for Federated
  Learning}.
\newblock \bibinfo{journal}{\emph{arXiv preprint arXiv:2003.08673}}
  (\bibinfo{year}{2020}).
\newblock


\bibitem[\protect\citeauthoryear{Lee, Chirkov, Ignasheva, Pisarchyk, Shieh,
  Riccardi, Sarokin, Kulik, and Grundmann}{Lee et~al\mbox{.}}{2019}]%
        {lee2019device}
\bibfield{author}{\bibinfo{person}{Juhyun Lee}, \bibinfo{person}{Nikolay
  Chirkov}, \bibinfo{person}{Ekaterina Ignasheva}, \bibinfo{person}{Yury
  Pisarchyk}, \bibinfo{person}{Mogan Shieh}, \bibinfo{person}{Fabio Riccardi},
  \bibinfo{person}{Raman Sarokin}, \bibinfo{person}{Andrei Kulik}, {and}
  \bibinfo{person}{Matthias Grundmann}.} \bibinfo{year}{2019}\natexlab{}.
\newblock \showarticletitle{On-device neural net inference with mobile gpus}.
\newblock \bibinfo{journal}{\emph{arXiv preprint arXiv:1907.01989}}
  (\bibinfo{year}{2019}).
\newblock


\bibitem[\protect\citeauthoryear{Lin, Ren, Chen, Ren, Yu, Ma, Rijke, and
  Cheng}{Lin et~al\mbox{.}}{2020}]%
        {lin2020meta}
\bibfield{author}{\bibinfo{person}{Yujie Lin}, \bibinfo{person}{Pengjie Ren},
  \bibinfo{person}{Zhumin Chen}, \bibinfo{person}{Zhaochun Ren},
  \bibinfo{person}{Dongxiao Yu}, \bibinfo{person}{Jun Ma},
  \bibinfo{person}{Maarten~de Rijke}, {and} \bibinfo{person}{Xiuzhen Cheng}.}
  \bibinfo{year}{2020}\natexlab{}.
\newblock \showarticletitle{Meta Matrix Factorization for Federated Rating
  Predictions}. In \bibinfo{booktitle}{\emph{Proceedings of the 43rd
  International ACM SIGIR Conference on Research and Development in Information
  Retrieval}}. \bibinfo{pages}{981--990}.
\newblock


\bibitem[\protect\citeauthoryear{Lu, Shu, Tan, Liu, Zhou, Chen, and Pei}{Lu
  et~al\mbox{.}}{2019}]%
        {lu2019collaborative}
\bibfield{author}{\bibinfo{person}{Yan Lu}, \bibinfo{person}{Yuanchao Shu},
  \bibinfo{person}{Xu Tan}, \bibinfo{person}{Yunxin Liu},
  \bibinfo{person}{Mengyu Zhou}, \bibinfo{person}{Qi Chen}, {and}
  \bibinfo{person}{Dan Pei}.} \bibinfo{year}{2019}\natexlab{}.
\newblock \showarticletitle{Collaborative learning between cloud and end
  devices: an empirical study on location prediction}. In
  \bibinfo{booktitle}{\emph{Proceedings of the 4th ACM/IEEE Symposium on Edge
  Computing}}. \bibinfo{pages}{139--151}.
\newblock


\bibitem[\protect\citeauthoryear{Muhammad, Wang, O'Reilly-Morgan, Tragos,
  Smyth, Hurley, Geraci, and Lawlor}{Muhammad et~al\mbox{.}}{2020}]%
        {muhammad2020fedfast}
\bibfield{author}{\bibinfo{person}{Khalil Muhammad}, \bibinfo{person}{Qinqin
  Wang}, \bibinfo{person}{Diarmuid O'Reilly-Morgan}, \bibinfo{person}{Elias
  Tragos}, \bibinfo{person}{Barry Smyth}, \bibinfo{person}{Neil Hurley},
  \bibinfo{person}{James Geraci}, {and} \bibinfo{person}{Aonghus Lawlor}.}
  \bibinfo{year}{2020}\natexlab{}.
\newblock \showarticletitle{Fedfast: Going beyond average for faster training
  of federated recommender systems}. In \bibinfo{booktitle}{\emph{Proceedings
  of the 26th ACM SIGKDD International Conference on Knowledge Discovery \&
  Data Mining}}. \bibinfo{pages}{1234--1242}.
\newblock


\bibitem[\protect\citeauthoryear{Niu, Wu, Tang, Hua, Jia, Lv, Wu, and Chen}{Niu
  et~al\mbox{.}}{2020}]%
        {niu2020billion}
\bibfield{author}{\bibinfo{person}{Chaoyue Niu}, \bibinfo{person}{Fan Wu},
  \bibinfo{person}{Shaojie Tang}, \bibinfo{person}{Lifeng Hua},
  \bibinfo{person}{Rongfei Jia}, \bibinfo{person}{Chengfei Lv},
  \bibinfo{person}{Zhihua Wu}, {and} \bibinfo{person}{Guihai Chen}.}
  \bibinfo{year}{2020}\natexlab{}.
\newblock \showarticletitle{Billion-scale federated learning on mobile clients:
  A submodel design with tunable privacy}. In
  \bibinfo{booktitle}{\emph{Proceedings of the 26th Annual International
  Conference on Mobile Computing and Networking}}. \bibinfo{pages}{1--14}.
\newblock


\bibitem[\protect\citeauthoryear{Park and Tuzhilin}{Park and Tuzhilin}{2008}]%
        {park2008long}
\bibfield{author}{\bibinfo{person}{Yoon-Joo Park} {and}
  \bibinfo{person}{Alexander Tuzhilin}.} \bibinfo{year}{2008}\natexlab{}.
\newblock \showarticletitle{The long tail of recommender systems and how to
  leverage it}. In \bibinfo{booktitle}{\emph{Proceedings of the 2008 ACM
  conference on Recommender systems}}. \bibinfo{pages}{11--18}.
\newblock


\bibitem[\protect\citeauthoryear{Perc}{Perc}{2014}]%
        {perc2014matthew}
\bibfield{author}{\bibinfo{person}{Matja{\v{z}} Perc}.}
  \bibinfo{year}{2014}\natexlab{}.
\newblock \showarticletitle{The Matthew effect in empirical data}.
\newblock \bibinfo{journal}{\emph{Journal of The Royal Society Interface}}
  \bibinfo{volume}{11}, \bibinfo{number}{98} (\bibinfo{year}{2014}),
  \bibinfo{pages}{20140378}.
\newblock


\bibitem[\protect\citeauthoryear{Qi, Wu, Wu, Huang, and Xie}{Qi
  et~al\mbox{.}}{2020}]%
        {qi2020privacy}
\bibfield{author}{\bibinfo{person}{Tao Qi}, \bibinfo{person}{Fangzhao Wu},
  \bibinfo{person}{Chuhan Wu}, \bibinfo{person}{Yongfeng Huang}, {and}
  \bibinfo{person}{Xing Xie}.} \bibinfo{year}{2020}\natexlab{}.
\newblock \showarticletitle{Privacy-Preserving News Recommendation Model
  Learning}. In \bibinfo{booktitle}{\emph{Proceedings of the 2020 Conference on
  Empirical Methods in Natural Language Processing: Findings}}.
  \bibinfo{pages}{1423--1432}.
\newblock


\bibitem[\protect\citeauthoryear{Rao, Yao, Zhang, and Zhang}{Rao
  et~al\mbox{.}}{2016}]%
        {rao2016preference}
\bibfield{author}{\bibinfo{person}{Zhiwei Rao}, \bibinfo{person}{Jiangchao
  Yao}, \bibinfo{person}{Ya Zhang}, {and} \bibinfo{person}{Rui Zhang}.}
  \bibinfo{year}{2016}\natexlab{}.
\newblock \showarticletitle{Preference aware recommendation based on
  categorical information}. In \bibinfo{booktitle}{\emph{2016 15th IEEE
  International Conference on Machine Learning and Applications (ICMLA)}}.
  IEEE, \bibinfo{pages}{865--870}.
\newblock


\bibitem[\protect\citeauthoryear{Rendle}{Rendle}{2010}]%
        {rendle2010factorization}
\bibfield{author}{\bibinfo{person}{Steffen Rendle}.}
  \bibinfo{year}{2010}\natexlab{}.
\newblock \showarticletitle{Factorization machines}. In
  \bibinfo{booktitle}{\emph{2010 IEEE International Conference on Data
  Mining}}. IEEE, \bibinfo{pages}{995--1000}.
\newblock


\bibitem[\protect\citeauthoryear{Resnick and Varian}{Resnick and
  Varian}{1997}]%
        {resnick1997recommender}
\bibfield{author}{\bibinfo{person}{Paul Resnick} {and} \bibinfo{person}{Hal~R
  Varian}.} \bibinfo{year}{1997}\natexlab{}.
\newblock \showarticletitle{Recommender systems}.
\newblock \bibinfo{journal}{\emph{Commun. ACM}} \bibinfo{volume}{40},
  \bibinfo{number}{3} (\bibinfo{year}{1997}), \bibinfo{pages}{56--58}.
\newblock


\bibitem[\protect\citeauthoryear{Sarwar, Karypis, Konstan, and Riedl}{Sarwar
  et~al\mbox{.}}{2001}]%
        {sarwar2001item}
\bibfield{author}{\bibinfo{person}{Badrul Sarwar}, \bibinfo{person}{George
  Karypis}, \bibinfo{person}{Joseph Konstan}, {and} \bibinfo{person}{John
  Riedl}.} \bibinfo{year}{2001}\natexlab{}.
\newblock \showarticletitle{Item-based collaborative filtering recommendation
  algorithms}. In \bibinfo{booktitle}{\emph{Proceedings of the 10th
  international conference on World Wide Web}}. \bibinfo{pages}{285--295}.
\newblock


\bibitem[\protect\citeauthoryear{Satyanarayanan}{Satyanarayanan}{2017}]%
        {satyanarayanan2017emergence}
\bibfield{author}{\bibinfo{person}{Mahadev Satyanarayanan}.}
  \bibinfo{year}{2017}\natexlab{}.
\newblock \showarticletitle{The emergence of edge computing}.
\newblock \bibinfo{journal}{\emph{Computer}} \bibinfo{volume}{50},
  \bibinfo{number}{1} (\bibinfo{year}{2017}), \bibinfo{pages}{30--39}.
\newblock


\bibitem[\protect\citeauthoryear{Shani, Heckerman, and Brafman}{Shani
  et~al\mbox{.}}{2005}]%
        {shani2005mdp}
\bibfield{author}{\bibinfo{person}{Guy Shani}, \bibinfo{person}{David
  Heckerman}, {and} \bibinfo{person}{Ronen~I Brafman}.}
  \bibinfo{year}{2005}\natexlab{}.
\newblock \showarticletitle{An MDP-based recommender system}.
\newblock \bibinfo{journal}{\emph{Journal of Machine Learning Research}}
  \bibinfo{volume}{6}, \bibinfo{number}{Sep} (\bibinfo{year}{2005}),
  \bibinfo{pages}{1265--1295}.
\newblock


\bibitem[\protect\citeauthoryear{Stoica, Song, Popa, Patterson, Mahoney, Katz,
  Joseph, Jordan, Hellerstein, Gonzalez, et~al\mbox{.}}{Stoica
  et~al\mbox{.}}{2017}]%
        {stoica2017berkeley}
\bibfield{author}{\bibinfo{person}{Ion Stoica}, \bibinfo{person}{Dawn Song},
  \bibinfo{person}{Raluca~Ada Popa}, \bibinfo{person}{David Patterson},
  \bibinfo{person}{Michael~W Mahoney}, \bibinfo{person}{Randy Katz},
  \bibinfo{person}{Anthony~D Joseph}, \bibinfo{person}{Michael Jordan},
  \bibinfo{person}{Joseph~M Hellerstein}, \bibinfo{person}{Joseph~E Gonzalez},
  {et~al\mbox{.}}} \bibinfo{year}{2017}\natexlab{}.
\newblock \showarticletitle{A berkeley view of systems challenges for ai}.
\newblock \bibinfo{journal}{\emph{arXiv preprint arXiv:1712.05855}}
  (\bibinfo{year}{2017}).
\newblock


\bibitem[\protect\citeauthoryear{Sun, Yuan, Yang, Wei, Zhao, and Liu}{Sun
  et~al\mbox{.}}{2020}]%
        {sun2020generic}
\bibfield{author}{\bibinfo{person}{Yang Sun}, \bibinfo{person}{Fajie Yuan},
  \bibinfo{person}{Ming Yang}, \bibinfo{person}{Guoao Wei},
  \bibinfo{person}{Zhou Zhao}, {and} \bibinfo{person}{Duo Liu}.}
  \bibinfo{year}{2020}\natexlab{}.
\newblock \showarticletitle{A Generic Network Compression Framework for
  Sequential Recommender Systems}.
\newblock \bibinfo{journal}{\emph{SIGIR}} (\bibinfo{year}{2020}).
\newblock


\bibitem[\protect\citeauthoryear{Sundaramoorthy, Gudur, Moorthy, Bhandari, and
  Vijayaraghavan}{Sundaramoorthy et~al\mbox{.}}{2018}]%
        {sundaramoorthy2018harnet}
\bibfield{author}{\bibinfo{person}{Prahalathan Sundaramoorthy},
  \bibinfo{person}{Gautham~Krishna Gudur}, \bibinfo{person}{Manav~Rajiv
  Moorthy}, \bibinfo{person}{R~Nidhi Bhandari}, {and} \bibinfo{person}{Vineeth
  Vijayaraghavan}.} \bibinfo{year}{2018}\natexlab{}.
\newblock \showarticletitle{Harnet: Towards on-device incremental learning
  using deep ensembles on constrained devices}. In
  \bibinfo{booktitle}{\emph{Proceedings of the 2nd International Workshop on
  Embedded and Mobile Deep Learning}}. \bibinfo{pages}{31--36}.
\newblock


\bibitem[\protect\citeauthoryear{Tan, Zhang, Yao, Liu, Zhou, Yang, and Hu}{Tan
  et~al\mbox{.}}{2021}]%
        {tan2021sparse}
\bibfield{author}{\bibinfo{person}{Qiaoyu Tan}, \bibinfo{person}{Jianwei
  Zhang}, \bibinfo{person}{Jiangchao Yao}, \bibinfo{person}{Ninghao Liu},
  \bibinfo{person}{Jingren Zhou}, \bibinfo{person}{Hongxia Yang}, {and}
  \bibinfo{person}{Xia Hu}.} \bibinfo{year}{2021}\natexlab{}.
\newblock \showarticletitle{Sparse-interest network for sequential
  recommendation}. In \bibinfo{booktitle}{\emph{Proceedings of the 14th ACM
  International Conference on Web Search and Data Mining}}.
  \bibinfo{pages}{598--606}.
\newblock


\bibitem[\protect\citeauthoryear{Yang, Liu, Chen, and Tong}{Yang
  et~al\mbox{.}}{2019}]%
        {yang2019federated}
\bibfield{author}{\bibinfo{person}{Qiang Yang}, \bibinfo{person}{Yang Liu},
  \bibinfo{person}{Tianjian Chen}, {and} \bibinfo{person}{Yongxin Tong}.}
  \bibinfo{year}{2019}\natexlab{}.
\newblock \showarticletitle{Federated machine learning: Concept and
  applications}.
\newblock \bibinfo{journal}{\emph{ACM Transactions on Intelligent Systems and
  Technology (TIST)}} \bibinfo{volume}{10}, \bibinfo{number}{2}
  (\bibinfo{year}{2019}), \bibinfo{pages}{1--19}.
\newblock


\bibitem[\protect\citeauthoryear{Yao, Zhang, Tsang, and Sun}{Yao
  et~al\mbox{.}}{2017}]%
        {yao2017discovering}
\bibfield{author}{\bibinfo{person}{Jiangchao Yao}, \bibinfo{person}{Ya Zhang},
  \bibinfo{person}{Ivor Tsang}, {and} \bibinfo{person}{Jun Sun}.}
  \bibinfo{year}{2017}\natexlab{}.
\newblock \showarticletitle{Discovering user interests from social images}. In
  \bibinfo{booktitle}{\emph{International Conference on Multimedia Modeling}}.
  Springer, \bibinfo{pages}{160--172}.
\newblock


\bibitem[\protect\citeauthoryear{Yuan, He, Karatzoglou, and Zhang}{Yuan
  et~al\mbox{.}}{2020}]%
        {yuan2020parameter}
\bibfield{author}{\bibinfo{person}{Fajie Yuan}, \bibinfo{person}{Xiangnan He},
  \bibinfo{person}{Alexandros Karatzoglou}, {and} \bibinfo{person}{Liguang
  Zhang}.} \bibinfo{year}{2020}\natexlab{}.
\newblock \showarticletitle{Parameter-efficient transfer from sequential
  behaviors for user modeling and recommendation}. In
  \bibinfo{booktitle}{\emph{Proceedings of the 43rd International ACM SIGIR
  Conference on Research and Development in Information Retrieval}}.
  \bibinfo{pages}{1469--1478}.
\newblock


\bibitem[\protect\citeauthoryear{Zhang, Yao, Zhao, Chua, and Wu}{Zhang
  et~al\mbox{.}}{2021}]%
        {zhang2021cause}
\bibfield{author}{\bibinfo{person}{Shengyu Zhang}, \bibinfo{person}{Dong Yao},
  \bibinfo{person}{Zhou Zhao}, \bibinfo{person}{Tat-Seng Chua}, {and}
  \bibinfo{person}{Fei Wu}.} \bibinfo{year}{2021}\natexlab{}.
\newblock \showarticletitle{CauseRec: Counterfactual User Sequence Synthesis
  for Sequential Recommendation}. In \bibinfo{booktitle}{\emph{Proceedings of
  the 44th International ACM SIGIR Conference on Research and Development in
  Information Retrieval}}.
\newblock


\bibitem[\protect\citeauthoryear{Zhao and Shang}{Zhao and Shang}{2010}]%
        {zhao2010user}
\bibfield{author}{\bibinfo{person}{Zhi-Dan Zhao} {and}
  \bibinfo{person}{Ming-Sheng Shang}.} \bibinfo{year}{2010}\natexlab{}.
\newblock \showarticletitle{User-based collaborative-filtering recommendation
  algorithms on hadoop}. In \bibinfo{booktitle}{\emph{2010 Third International
  Conference on Knowledge Discovery and Data Mining}}. IEEE,
  \bibinfo{pages}{478--481}.
\newblock


\bibitem[\protect\citeauthoryear{Zhou, Zhu, Song, Fan, Zhu, Ma, Yan, Jin, Li,
  and Gai}{Zhou et~al\mbox{.}}{2018}]%
        {zhou2018deep}
\bibfield{author}{\bibinfo{person}{Guorui Zhou}, \bibinfo{person}{Xiaoqiang
  Zhu}, \bibinfo{person}{Chenru Song}, \bibinfo{person}{Ying Fan},
  \bibinfo{person}{Han Zhu}, \bibinfo{person}{Xiao Ma},
  \bibinfo{person}{Yanghui Yan}, \bibinfo{person}{Junqi Jin},
  \bibinfo{person}{Han Li}, {and} \bibinfo{person}{Kun Gai}.}
  \bibinfo{year}{2018}\natexlab{}.
\newblock \showarticletitle{Deep interest network for click-through rate
  prediction}. In \bibinfo{booktitle}{\emph{Proceedings of the 24th ACM SIGKDD
  International Conference on Knowledge Discovery \& Data Mining}}.
  \bibinfo{pages}{1059--1068}.
\newblock


\end{thebibliography}

\clearpage
\appendix

\section{Reproducibility Details}
\subsection{Dataset Generation}
\begin{figure}[!h]
    \centering
    \includegraphics[width=0.4\textwidth]{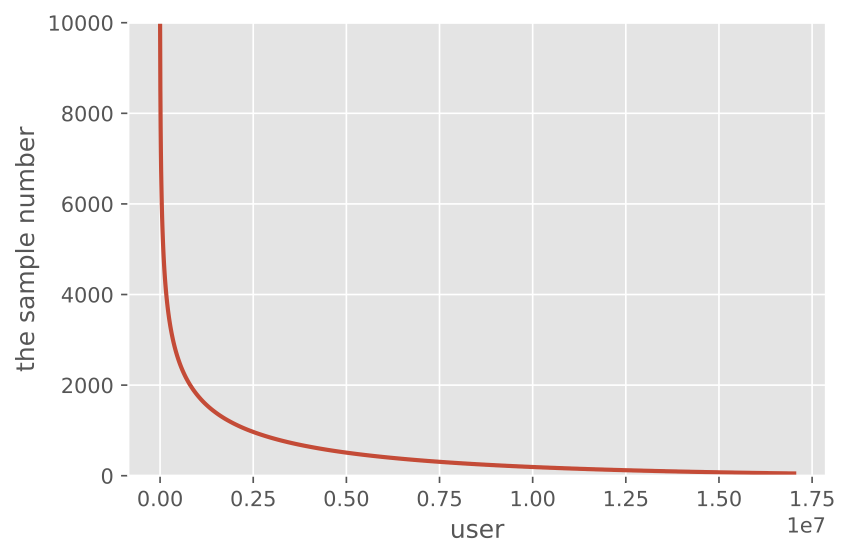}
    \caption{The Taobao sample number \textit{w.r.t.} the user index.}
    \label{fig:user_sample}
\end{figure}

\textbf{Amazon}\footnote{\url{https://nijianmo.github.io/amazon/index.html}} and \textbf{Movielens-1M}\footnote{\url{https://grouplens.org/datasets/movielens/1m/}} are public datasets containing user-item interactions. We collect \textbf{Taobao} dataset from Mobile Taobao App\footnote{\url{https://market.m.taobao.com/app/fdilab/download-page/main/index.html}}, containing historical behaviors of randomly sampled users in one month. We show the line plot of sorted number of user behaviors in descending order in Figure~\ref{fig:user_sample}. One can observe the long-tail effect of data distribution, which indicates the necessity of personalized model.

We preprocess the datasets to guarantee that users and items have been interacted at least 8 times in Amazon dataset, 20 times in Movielens-1M dataset and 50 times in Taobao dataset.

To train one-round DCCL, each dataset is divided into three disjoint parts according to the log timestamps, including pre-training phase, device-cloud collaborative learning phase and test phase. Pre-training phase is used to learn the initial cloud model $f$ and globally shared parameter basis $\Theta$. Then, \emph{MetaPatch} and \emph{MoMoDistill} are performed successively on device-cloud collaborative learning phase. Finally the model is evaluated on the test phase dataset. For multi-round DCCL in RQ3, the device-cloud collaborative learning phase is further equally divided into multiple parts, each of which corresponding to one round.

\subsection{Parameter Settings}
Table~\ref{tab:params} shows the hyper-parameter setting of DCCL method for three datasets.

\subsection{Experiment Environment}
All experiments are conducted on workstations equipped with GPUs (Tesla V100). All baselines and DCCL method are implemented with TensorFlow. Software versions of Python and TensorFlow are 2.7 and 1.12 respectively.

% \usepackage{booktabs}
% \usepackage{multirow}
\begin{table}[!b]
\centering
\caption{Hyper parameters of DCCL in Amazon, Movielens, Taobao dataset}
\label{tab:params}
\begin{tabular}{@{}lll@{}}
\toprule
Dataset                     & Parameters                                     & Setting         \\ \midrule
\multirow{11}{*}{Amazon}    & user embedding dimension                       & 15              \\ \cmidrule(l){2-3} 
                            & item embedding dimension                       & 10              \\ \cmidrule(l){2-3} 
                            & brand embedding dimension                      & 10              \\ \cmidrule(l){2-3} 
                            & category embedding dimension                   & 10              \\ \cmidrule(l){2-3} 
                            & learning rate          & 1e-3 \\ \cmidrule(l){2-3} 
                            & batch size                                     & 512             \\ \cmidrule(l){2-3} 
                            & num\_attention\_layers                         & 2               \\ \cmidrule(l){2-3} 
                            & attention dimension(Q,K,V)                     & 32              \\ \cmidrule(l){2-3} 
                            & feed\_forward\_layer\_dimension                & 32,16           \\ \cmidrule(l){2-3} 
                            & patch dimension                                & 32,16,32           \\ \cmidrule(l){2-3} 
                            & optimizer                                      & Adam \\ \cmidrule(l){2-3} 
                            & $\beta$ & 0.01
                            \\ \midrule
\multirow{11}{*}{Movielens-1M} & user embedding                                 & 8               \\ \cmidrule(l){2-3} 
                            & item embedding                                 & 8               \\ \cmidrule(l){2-3} 
                            & learning rate                                  & 1e-3            \\ \cmidrule(l){2-3} 
                            & optimizer                                      & Adam            \\ \cmidrule(l){2-3} 
                            & batch size                                     & 256             \\ \cmidrule(l){2-3} 
                            & encoder layer                                  & 2               \\ \cmidrule(l){2-3} 
                            & encoder size                                   & 32              \\ \cmidrule(l){2-3} 
                            & attention dimension(Q,K,V)                     & 32              \\ \cmidrule(l){2-3} 
                            & classifer dimension                            & 32              \\ \cmidrule(l){2-3} 
                            & patch dimension                                & 32 ,16, 32      \\ \cmidrule(l){2-3} 
                            & User Profile Dimension & 8 \\ \cmidrule(l){2-3} 
                            & $\beta$ & 0.01               \\ \midrule
\multirow{11}{*}{Taobao}    & user embedding                                 & 8               \\ \cmidrule(l){2-3} 
                            & item embedding                                 & 8               \\ \cmidrule(l){2-3} 
                            & learning rate                                  & 1e-3            \\ \cmidrule(l){2-3} 
                            & optimizer                                      & Adam            \\ \cmidrule(l){2-3} 
                            & batch size                                     & 1024            \\ \cmidrule(l){2-3} 
                            & encoder layer                                  & 2               \\ \cmidrule(l){2-3} 
                            & encoder size                                   & 32              \\ \cmidrule(l){2-3} 
                            & attention dimension(Q,K,V)                     & 32              \\ \cmidrule(l){2-3} 
                            & classifer dimension                            & 32              \\ \cmidrule(l){2-3} 
                            & patch dimension                                & 32 ,16, 32      \\ \cmidrule(l){2-3} 
                            & User Profile Dimension & 8   \\ \cmidrule(l){2-3} 
                            & $\beta$ & 0.01            \\ \bottomrule
\end{tabular}
\end{table}

\end{document}